\documentclass[sigconf,nonacm]{acmart}
\setcopyright{none}
\renewcommand\footnotetextcopyrightpermission[1]{}
\usepackage{amsfonts,bm}

\def\eqref#1{equation~\ref{#1}}

\def\1{\bm{1}}

\DeclareMathAlphabet{\mathsfit}{\encodingdefault}{\sfdefault}{m}{sl}
\SetMathAlphabet{\mathsfit}{bold}{\encodingdefault}{\sfdefault}{bx}{n}

\usepackage{booktabs}
\usepackage{microtype} 
\usepackage{graphicx}
\usepackage{enumitem}
\usepackage{multirow}
\usepackage{amssymb}
\usepackage{tikz}
\usetikzlibrary{positioning,arrows.meta,calc,shapes.geometric,fit,backgrounds}
\usepackage[most]{tcolorbox}
\definecolor{promptblue}{RGB}{30,70,200}
\definecolor{promptred}{RGB}{160,20,70}
\definecolor{promptgreen}{RGB}{20,115,60}
\newtcolorbox{promptbox}[1]{colback=gray!7, colframe=black, colbacktitle=black,
  coltitle=white, fonttitle=\bfseries, title={#1}, boxrule=0.5pt, arc=2.5pt,
  left=5pt, right=5pt, top=4pt, bottom=4pt, breakable}

\title{Agentic Graph Token Reasoning}

\author{Zhuoyi Peng}
\affiliation{%
  \institution{The Hong Kong University of Science and Technology}
  \country{}
}
\email{zpengaf@connect.ust.hk}

\author{Yi Yang}
\affiliation{%
  \institution{The Hong Kong University of Science and Technology}
  \country{}
}
\email{imyiyang@ust.hk}

\begin{document}

\begin{abstract}
Graphs model relational data throughout science and industry, from citation
networks and protein interactions to product co-purchase graphs. Because the
nodes of many such graphs carry rich text, a growing line of work applies large
language models (LLMs) to graph analysis. The most graph-native of these
methods use \emph{graph tokens}: a graph encoder compresses a graph view---a
subgraph such as a node, its $k$-hop neighbourhood, or a cluster---into a short
block of continuous tokens that jointly encodes node attributes and topology
and is read directly by the model. Existing methods, however, use graph tokens
in a static single-shot manner: they encode one predefined graph view before the model
has even seen the target and never revise it, leaving the model's step-by-step
reasoning ability unused for a task. We introduce
\emph{agentic graph token reasoning}, which recasts graph tokenization as part
of the reasoning process itself. At each step, the model chooses which graph
view to encode and at what granularity; a graph encoder is invoked on demand to
materialise the corresponding graph tokens; and the resulting block is spliced
into the running context. The model thus reasons step by step in the graph
token space, and the tokens it reads are trajectory-dependent. We realise this
with a three-stage training pipeline: (i) self-supervised tasks that teach the
model to read heterogeneous graph tokens, (ii) a token-robust trajectory stage
with a graph-token consistency regulariser, and (iii) preference optimisation
that rewards trajectories in which the graph-token evidence and the node-text
evidence agree, incentivising the model to ground its predictions in the
consistency between the two. Across extensive evaluations spanning seven graph
domains, our models outperform a broad set of baselines by a large margin and
transfer zero-shot to unseen domains without any per-target fine-tuning.
Further analysis shows that our agentic approach substantially reduces errors
on the most challenging samples. More broadly, this work pushes LLM-based
graph analysis from static graph-token encoders towards a graph-native agent
paradigm.
\end{abstract}

 
\maketitle
 
\section{Introduction}
 
Graphs are a universal representation for relational data, and in many domains
their nodes carry rich text, a paper and its abstract, a product and its
description, a protein and its functional annotation. A promising line of recent
progress uses large language models to analyse such graphs: one asks a question
about the graph (what is this paper's field? do these two proteins interact?), and the model returns the answer. The challenge is that a language model
is trained purely on language: aligning graph data with its language space, so that
the model can understand a graph at all, is difficult. The technique that has
emerged for this is the \emph{graph token}. Methods such as
LLaGA~\cite{llaga2024}, TEA-GLM~\cite{teaglm2024}, GraphGPT~\cite{graphgpt2024},
GraphTranslator~\cite{graphtranslator2024}, and GOFA~\cite{gofa2025}
run a graph encoder over a graph view\footnote{A \emph{graph
view} is a subgraph of the full graph, or the full graph itself, including but not limited to the target node, its $k$-hop neighbourhood, or a cluster, forming a graph region for the language model to process.} around the target and compress it
into a short, fixed-length block of continuous tokens that the model reads
alongside the question. Their success comes from exactly this compression: the
graph is turned into tokens a language model can understand, and the block stays
the same length however large the encoded graph view.
 
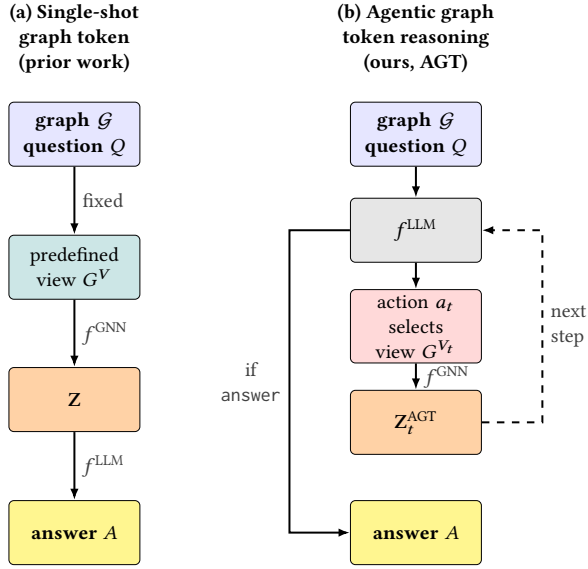
\begin{figure}[t]
\centering
\resizebox{\columnwidth}{!}{%
\begin{tikzpicture}[
  font=\small,
  b/.style={draw, rounded corners=2pt, minimum height=9mm, text width=17mm, align=center, inner sep=2pt},
  inp/.style={b, fill=blue!10, font=\small\bfseries},
  viewb/.style={b, fill=teal!20},
  zb/.style={b, fill=orange!35, font=\small\bfseries},
  actb/.style={b, fill=red!15},
  llmb/.style={b, fill=gray!20, font=\small\bfseries},
  ansb/.style={b, fill=yellow!55, font=\small\bfseries},
  arr/.style={-{Latex[length=1.6mm]}, thick},
  arrb/.style={-{Latex[length=1.6mm]}, thick, dashed},
  ttl/.style={font=\small\bfseries, align=center},
  lab/.style={font=\small, text=black!75},
]
\path (-4.1,0);
\node[ttl] at (-2.4,1.35) {(a) Single-shot\\graph token\\(prior work)};
\node[inp]   (a1) at (-2.4,0)    {graph $\mathcal{G}$\\question $Q$};
\node[viewb] (a2) at (-2.4,-1.87) {predefined\\view $G^{V}$};
\node[zb]    (a3) at (-2.4,-3.73) {$\mathbf{Z}$};
\node[ansb]  (a4) at (-2.4,-5.60) {answer $A$};
\draw[arr] (a1) -- node[right,lab]{fixed} (a2);
\draw[arr] (a2) -- node[right,lab]{$f^{\mathrm{GNN}}$} (a3);
\draw[arr] (a3) -- node[right,lab]{$f^{\mathrm{LLM}}$} (a4);
 
\node[ttl] at (2.4,1.35) {(b) Agentic graph\\token reasoning\\(ours, AGT)};
\node[inp]  (b1) at (2.4,0)    {graph $\mathcal{G}$\\question $Q$};
\node[llmb] (b2) at (2.4,-1.35) {$f^{\mathrm{LLM}}$};
\node[actb] (b3) at (2.4,-2.70) {action $a_t$\\selects view $G^{V_t}$};
\node[zb]   (b4) at (2.4,-4.05) {$\mathbf{Z}^{\mathrm{AGT}}_t$};
\node[ansb] (b5) at (2.4,-5.60) {answer $A$};
\draw[arr] (b1) -- (b2);
\draw[arr] (b2) -- (b3);
\draw[arr] (b3) -- node[right,lab]{$f^{\mathrm{GNN}}$} (b4);
\draw[arrb] (b4.east) -- ++(0.85,0) |- node[right,lab,pos=0.25,align=left]{next\\step} (b2.east);
\draw[arr] (b2.west) -- ++(-0.85,0) |- node[left,lab,pos=0.25,align=center]{if\\\texttt{answer}} (b5.west);
\end{tikzpicture}%
}
\caption{Single-shot vs.\ AGT, from the same inputs: a graph
$\mathcal{G}$ and a question $Q$. (a) Prior graph-token methods encode one
predefined view into a graph token $\mathbf{Z}$ and answer in a single pass. (b) Ours
lets the model emit an action at each step that calls a graph view to encode; the
encoder returns an agentic graph-token block that is spliced back into the
context, and the loop repeats until the terminating \texttt{answer} action. A
fraud check, for example, may read the account's own text, then its transaction
partners, the surrounding community, and look-alike accounts before answering.}
\label{fig:compare}
\end{figure}

These methods, however, read the graph using a predefined, static graph-token
pattern. Given a target, the graph view
to encode, typically the anchor's $k$-hop neighbourhood or a single global token, is chosen in advance, encoded before the model has even seen the question,
and never revised; the model must then produce its answer in a single forward pass
(Figure~\ref{fig:compare}a). Everything therefore rests on the quality of that one
chosen graph view: it must be neither too narrow to contain the evidence nor so wide
that the signal is diluted, and it is committed before the question is known. This
raises a basic concern: how can one assume that a predefined, static graph token is
sufficient for the language model to read and derive the final answer in one shot?
 
Meanwhile, language models have developed a capability that speaks directly to this
problem: \emph{agentic reasoning}. Rather than answering from whatever happens to be
in the prompt, a model can now decide what it still needs, act to obtain it, and
repeat, a behaviour typically instilled through large-scale reinforcement
learning~\cite{grpo2024,o1_2024,deepseekr1_2025}. The benefit of such iteration is that the
context is enriched round by round, each step adding evidence chosen in light of what the previous steps revealed, until it is sufficient to derive the answer, so how
much evidence is gathered adapts to how hard the instance is instead of being fixed
beforehand. Graph analysis is a natural fit for this, since the evidence a target
needs (its own attributes, its neighbours, the community around it, look-alike nodes elsewhere) is rarely apparent before one starts looking. Yet current
graph-token methods lag far behind this paradigm: they commit to one graph view up
front and answer in a single pass, so the model never gets to ask for the evidence it
turns out to need. As a result, the capability of language models for graph analysis
is far from fully unleashed. Closing this gap calls for a multi-step formulation in
which reasoning proceeds over several rounds, each represented by graph
tokens, and
leads to our research question:
 
\smallskip
\noindent\emph{How can a language model perform step-by-step agentic reasoning via graph
tokens?}
\smallskip

We address this question with \emph{Agentic Graph Token} (AGT) reasoning
(Figure~\ref{fig:compare}b), which turns graph tokenization from a
pre-processing step into part of the reasoning process. Given a graph and a task
question, the model reasons one step at a time: it emits a discrete action that
names a scope and a granularity, from the anchor node itself up to a $k$-hop subgraph, a
retrieved set of semantically similar nodes, or a whole cluster, and a graph encoder is
invoked on demand to encode that graph view into a fixed-length block of graph
tokens, and the block is spliced into the running context before the next step.
The model thereby grows a reasoning trajectory\footnote{A \emph{trajectory} is the
sequence of interleaved reasoning steps and actions a model produces while solving a
query, ending in its answer.} adaptively, each step chosen in light of the reasoning
so far, and adaptively terminates to output the task answer. Reasoning thus
unfolds in the graph token space rather than in language, in the spirit of
reasoning in a continuous latent space~\cite{coconut2024}.
 
Eliciting this behaviour is non-trivial, chiefly because a model given both
prompt text and graph tokens tends to lean on the text and ignore the tokens. We
therefore train the model in three stages, each contributing a component that, to
our knowledge, is new to graph-token learning: (1) \emph{learning to read graph
tokens}, self-supervised pre-training that aligns the encoder with the language
model so it can read the varied token blocks that different actions produce; (2)
\emph{incentivising reasoning trajectories}, trajectory SFT with a graph-token
consistency term so the reasoning is driven by token content rather than brittle
cues; and (3) \emph{graph-text consistency preference}, IPO that prefers rollouts
which stay consistent with the graph over ones a corrupted anchor text throws off,
pushing the policy to ground its answers in the graph.
 
\begin{figure*}[t]
\centering
\includegraphics[width=0.9\textwidth]{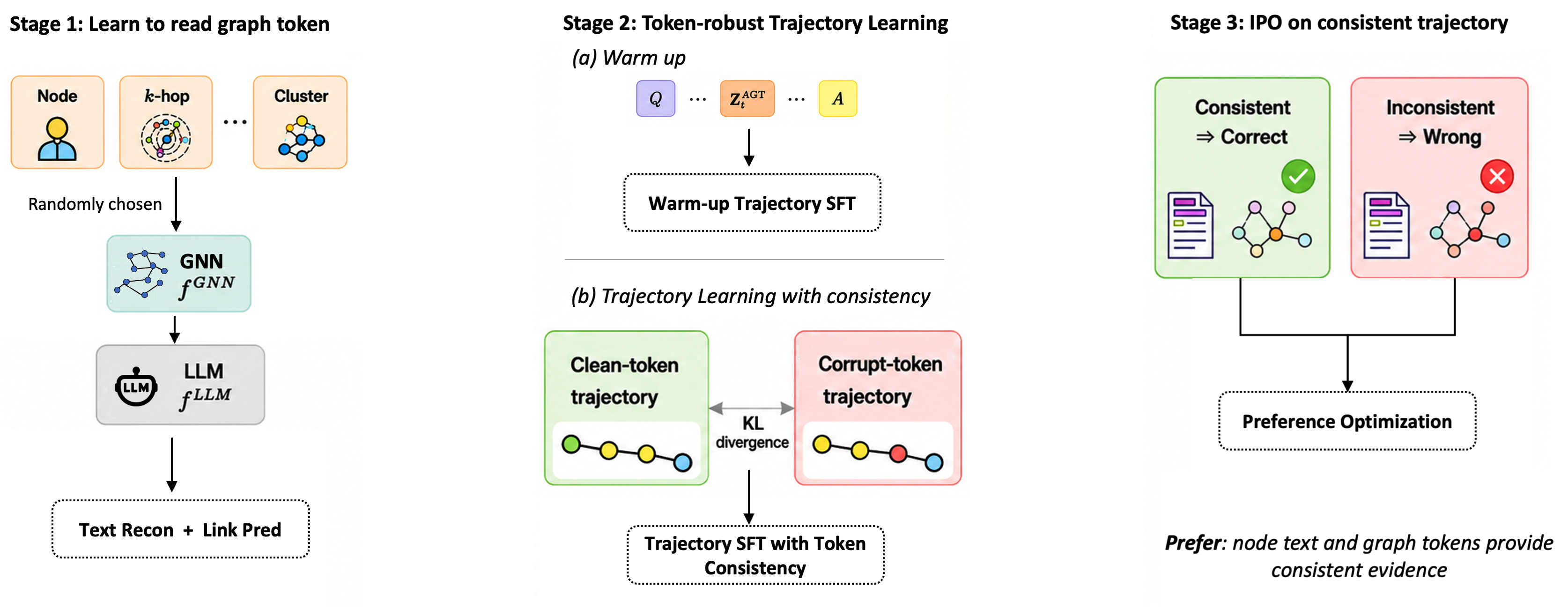}
\caption{Overview of Agentic Graph Token (AGT) reasoning. Stage~1 grounds the block with two
self-supervised reading tasks; Stage~2 fits agentic trajectories with a
consistency term over perturbed graph tokens; Stage~3 optimises preference pairs in
which a text-consistent rollout is correct and a text-corrupted, graph-inconsistent
one is wrong, teaching model to prefer the reasoning path where graph token and node text agree. LLM and GNN are jointly optimised in all stages.}
\label{fig:overview}
\end{figure*}
 
\paragraph{Contributions} Our contributions are threefold.
\begin{itemize}[leftmargin=1.2em,itemsep=3pt,topsep=3pt]
\item \textbf{An agentic graph-token reasoning framework.} We recast graph
tokenization as part of reasoning: the model interleaves its reasoning with
on-demand graph-token materialisations in a single sequence, moving graph learning
from a static, single-shot paradigm to an agentic one whose intermediate evidence
is graph-native rather than verbalised text.
\item \textbf{A three-stage training recipe.} We contribute one targeted design per
stage, teaching the model to read graph tokens, keeping its reading grounded in
their content, and rewarding trajectories where the graph and text evidence agree.
\item \textbf{Strong empirical results.} Across seven graph domains our model
outperforms comparably sized graph-token and agentic baselines on node
classification and link prediction, and transfers zero-shot to targets the checkpoint was
never trained on.
\end{itemize}
 
This work pushes LLM-based
graph analysis from static graph-token encoders towards a graph-native agent
paradigm.
 
\section{Related Work}
 
\subsection{Graph learning with large language models}
A large body of work adapts large language models to text-attributed graphs, and
the approaches differ mainly in how the graph is presented to the model.
\emph{Verbalisation} methods render local structure as natural-language text that
the model reads in the prompt: they serialise a node's neighbourhood into a
description~\cite{instructglm2023,graphtext2023,gpt4graph2023}, use an LLM to
generate explanations or augmented features for a downstream
model~\cite{tape2024,gaugllm2024}, or select the most informative neighbours to
describe~\cite{neighborselect2024}; broad empirical studies compare such text
encodings~\cite{chen2024explore}. Such text encodes topology poorly and inflates
the context. \emph{Graph-token} methods instead learn a graph encoder that
compresses a graph view into structure-aware soft tokens injected into the prompt:
LLaGA~\cite{llaga2024}, GOFA~\cite{gofa2025}, TEA-GLM~\cite{teaglm2024},
GraphGPT~\cite{graphgpt2024}, GraphTranslator~\cite{graphtranslator2024}, and
GraphPrompter~\cite{graphprompter2024} follow this route, typically attaching a GNN
encoder to an open LLM backbone such as LLaMA or Mistral~\cite{llama2023,mistral2023}.
A training-free variant instead performs graph in-context learning without any
tuning~\cite{graphicl2025}, and complementary analyses ask when and why structural
prompts help at all~\cite{huang2023graph}. These
representations are graph-native, but the encoded graph view is a fixed template chosen
at architecture time, encoded once before the model reads the target, and never
revised, so the model cannot request a different graph view as it reasons. A third,
recent \emph{agentic} line lets the model gather evidence over several steps:
AgentGL~\cite{agentgl2026} trains an RL policy that issues text queries and reads
back text snippets, and GraphCoT~\cite{graphcot2024} and
GraphSearch~\cite{graphsearch2026} likewise operate over text renderings of the
graph; relatedly, Graph-of-Thoughts~\cite{got2024} organises the model's own
textual thoughts as a graph rather than analysing input graph data. None of
these methods ever re-encodes graph tokens. In
short, existing work is either agentic but text-based, or graph-token-based but
static. To our knowledge, ours is the first agentic graph-token reasoning
framework, in which each reasoning step is itself an on-demand graph-token
materialisation, uniting the graph-native representation of graph-token methods
with the adaptivity of agentic ones. This also makes it differ fundamentally from
any text-based reasoning method, such as chain-of-thought
prompting~\cite{cot2022} or AgentGL~\cite{agentgl2026}: each reasoning step
carries graph-topology and node-attribute information in the token block itself,
which a purely textual step can convey only clumsily, if at all.
 
\subsection{Reasoning in large language models}
A parallel line of work improves how language models reason. Chain-of-thought
prompting~\cite{cot2022} shows that generating intermediate steps in language
before the final answer substantially improves multi-step problem solving, and a
large family of methods builds on this idea of explicit step-by-step reasoning.
Such reasoning, however, is carried out entirely in language. A growing line
argues that language is not always the best medium and moves computation off the
token stream: pause tokens add latent compute before the model
answers~\cite{pause2024}, implicit chain-of-thought internalises the reasoning into
hidden states~\cite{icot2024}, and Coconut~\cite{coconut2024} lets a
model reason in a continuous latent space, showing that a latent state can carry
information that a verbalised step cannot. Our setting inherits exactly this
tension, but in a graph context, where each reasoning step should convey
structural evidence about the graph. Graph tokens are a natural latent medium for
such steps, yet \textit{how to represent a graph reasoning step efficiently and
effectively} remains an open question, which this work sets out to address.
 
\section{Preliminaries}
\label{sec:prelim}
 
We study learning on a \emph{text-attributed graph}
(TAG) $\mathcal{G}=(\mathcal{V},\mathcal{E},\mathcal{T})$ with node set
$\mathcal{V}$, edge set $\mathcal{E}$, and node texts
$\mathcal{T}=\{t_v\mid v\in\mathcal{V}\}$. A task instance is a question $Q$ about
the graph, with a ground-truth label $y\in\mathcal{Y}$, and the model returns an
answer $A\in\mathcal{Y}$. The question may concern a node, a node pair, or more
generally any graph-defined object whose label depends on the surrounding
structure. Unlike a classifier that maps a fixed precomputed feature to its
prediction, our model forms $A$ from evidence drawn from $\mathcal{G}$ at inference
time: the target's own text, its neighbours' texts, and the surrounding structure.
 
\paragraph{Graph tokens via a GNN} The interface to the graph is a graph neural
network $f^{\mathrm{GNN}}$ that maps a graph view $G^{V}$, a subgraph of $\mathcal{G}$
(or $\mathcal{G}$ itself) such as a node, a $k$-hop neighbourhood, or a cluster, to a fixed-size block of soft
tokens
\begin{equation}  \mathbf{Z}=f^{\mathrm{GNN}}(G^{V}).
\end{equation}The block has the same length for every graph view and lies directly in the LLM's
embedding space, so any graph view, from a single node to a large neighbourhood, is
summarised by a constant-length set of tokens the language model can read.
 
\paragraph{Reading graph tokens with an LLM} A pretrained large language model
$f^{\mathrm{LLM}}$ is the backbone: it consumes accompanying text together with one
or more graph-token blocks $\mathbf{Z}=f^{\mathrm{GNN}}(G^{V})$ and produces the
answer. In the simplest, single-shot form used by prior graph-LLMs, a fixed graph view
$G^{V}$ is encoded once and read alongside the question $Q$,
\begin{equation}
  A \;=\; f^{\mathrm{LLM}}\!\left(Q,\ \mathbf{Z}\right),
  \qquad \mathbf{Z}=f^{\mathrm{GNN}}(G^{V}),
  \label{eq:singleshot}
\end{equation}
in one forward pass. Here the graph view is chosen at architecture time, before the
model has seen $Q$, and never revised. Our departure is exactly in \emph{when} and
\emph{which} blocks are read: rather than a single prepended block, the model
requests blocks step by step during generation, so the graph tokens it reads become
trajectory-dependent. Section~\ref{sec:method} develops this and contrasts it
directly with Eq.~\eqref{eq:singleshot}.
 
\section{Methodology}
\label{sec:method}
 
\textbf{Graph analysis needs step-by-step reasoning, and each step should be a graph token.}
Consider predicting a protein's function. For a well-characterised protein its own
annotation settles it; for a less studied one the function emerges only from its
direct interaction partners; and for a poorly annotated protein the answer appears
only once the larger pathway or complex it belongs to comes into view. The
informative graph view is thus instance-dependent and is not known in advance, so the
model has to gather evidence over several steps rather than from one fixed view. A
single graph-token block, fixed before the model has seen the target, cannot match
this: it over-encodes easy targets, starves hard ones, and cannot be revised once
decoding has begun.
Once reasoning is step by step, the evidence each step needs is inherently
\emph{structural}: a neighbourhood, a pathway, a cluster. A graph token encodes such
structure natively and compactly, packing node attributes and topology into a
fixed-size block, whereas verbalising the same graph view would inflate the context and
bury the topology in text. This makes the graph token the right medium for a
reasoning step on a graph. We therefore let the model choose the graph view \emph{as it
reasons}, using the interface of Section~\ref{sec:prelim}, where a graph view $G^{V}$
encodes to a fixed-size block $\mathbf{Z}=f^{\mathrm{GNN}}(G^{V})$ in
$f^{\mathrm{LLM}}$'s
embedding space.
 
\textbf{From single-shot to step-by-step graph token reasoning.}
Prior graph-LLMs use the interface of Eq.~\eqref{eq:singleshot} once: a single
graph view is fixed, encoded, and read in one forward pass to produce the answer. We
instead unroll the computation over $T$ steps and produce the
answer only at the end. To begin with, the model is given the graph $\mathcal{G}$ and the question $Q$, and
the context holds the question alone, with no graph tokens yet:
\begin{equation}  \tau_0=Q .
\end{equation}At each step $t=1,\dots,T$, the model reads the context so far and
emits a single \emph{action} $a_t$ from the action space $\mathcal{A}$
(Table~\ref{tab:actions}), no natural language, only the action itself, which
names a graph view $G^{V_t}$, where $V_t=V(a_t)$ is determined by the action; a GNN
encoder $f^{\mathrm{GNN}}$ turns that graph view into the step's \emph{agentic
graph-token} (AGT) block, and the block is appended to the context:
\begin{align}
  a_t &\sim f^{\mathrm{LLM}}(\,\cdot\mid \tau_{t-1}), \\
  \mathbf{Z}^{\mathrm{AGT}}_t &= f^{\mathrm{GNN}}\!\big(G^{V_t}\big),\qquad V_t=V(a_t), \\
  \tau_t &= \big(\tau_{t-1},\, a_t,\, \mathbf{Z}^{\mathrm{AGT}}_t\big).
\end{align}
The block $\mathbf{Z}^{\mathrm{AGT}}_t$ thus depends on the action $a_t$ through the
graph view $G^{V_t}$ it names. What step $t{+}1$ conditions on is therefore the token
block $\mathbf{Z}^{\mathrm{AGT}}_t$ itself, not a verbalisation of $G^{V_t}$.
Crucially, $\mathbf{Z}^{\mathrm{AGT}}_t$ is then read at the next step, it is part
of the context $\tau_t$ that step $t{+}1$ conditions on, so the graph tokens
gathered so far steer which graph view is requested next. And because the graph view is
chosen anew every step, $\mathbf{Z}^{\mathrm{AGT}}_t$ can differ from round to
round: the model may zoom in, widen, or jump to a different part of the graph
rather than reusing one fixed block.
 
Specifically, if $a_t$ is the terminating action, the model outputs the final
answer $A$ and the loop ends. The number of steps $T$ and the graph views visited are
chosen by the model, so they vary with $Q$. The action set in
Table~\ref{tab:actions} is by no means a complete action space; we choose these
actions for their generality across graphs and tasks, and the interface is open, so
follow-up work can add its own actions without changing the rest of the method.
 
We give the model this behaviour with three training stages, each initialised from
the previous one: Stage~1 aligns the encoder's tokens with the language model's
embedding space so a block can be read at all; Stage~2 teaches the model to emit
graph-token actions and answer from what they return; and Stage~3 sharpens the
resulting policy through preference optimisation on its own rollouts. The three
training corpora are constructed as detailed in \textbf{Appendix~\ref{app:corpus}}. We first
fix the token interface and encoder, then develop each stage.
Figure~\ref{fig:overview} summarises both the inference-time rollout and the three
training stages.
 
\begin{table}[t]
\centering
\small
\setlength{\tabcolsep}{4pt}
\caption{The action space $\mathcal{A}$ used in this work. This is by no means an exhaustive set, but a
representative one: we choose these actions for their generality across graphs and
tasks.}
\label{tab:actions}
\begin{tabular}{ll}
\toprule
Action & Encoded scope / effect \\
\midrule
\verb|node_token|         & the anchor node $v$ itself \\
\verb|one_hop_token|   & $v$ and its 1-hop neighbours \\
\verb|two_hop_token|   & the 2-hop neighbourhood of $v$ \\
\verb|three_hop_token| & the 3-hop neighbourhood of $v$ \\
\verb|retrieval_token|           & top-$k$ cosine-similar nodes (star graph) \\
\verb|cluster_token|      & the cluster containing $v$ \\
\midrule
\verb|anchor_text|        & $v$'s verbatim text (no encoding) \\
\midrule
\verb|cosine_sim|        & cosine similarity (only for link prediction) \\
\midrule
\verb|answer|             & end the episode; emit answer $A$ \\
\bottomrule
\end{tabular}
\end{table}
 
\subsection{Stage 1: Learning to read graph tokens}
 
A graph-token block is a set of continuous vectors that match no word in the
language model's vocabulary, so out of the box the model has no idea what they mean.
Stage~1 grounds them: it teaches the model that a block $\mathbf{Z}^{\mathrm{AGT}}$
carries graph-derived evidence, node content and local structure, rather than
text. For each training anchor $v$ we build two
self-supervised tasks. \emph{Text reconstruction} feeds $v$'s \verb|node_token|
block and asks the model to reproduce $v$'s title and abstract, forcing the block
to encode content; \emph{masked link prediction} feeds $v$'s block together with a
candidate $w$ and asks \verb|yes|/\verb|no| whether $(v,w)$ is an edge (every edge
touching a validation or test node removed), forcing it to encode local structure.
The objective is a cross-entropy over the model's response tokens, with the
graph-token positions masked out,
\begin{equation}  \mathcal{L}_1 \;=\; -\!\sum_{t\in\mathcal{S}}\log
  f^{\mathrm{LLM}}\!\left(y_t\mid y_{<t},\mathbf{Z}^{\mathrm{AGT}}\right),
\end{equation}where $\mathcal{S}$ indexes the response tokens and $\mathbf{Z}^{\mathrm{AGT}}$ is an
input rather than a target. This objective updates the graph encoder
$f^{\mathrm{GNN}}$ and the language model $f^{\mathrm{LLM}}$ jointly, with the mpnet
text features and the graph topology held frozen. Unlike prior graph-token methods, which tie the encoder
to a single task-specific instruction-tuning objective, these two task-agnostic
objectives ground the block on its own content and structure, so one reader
transfers across the heterogeneous blocks that different actions later produce.
 
\subsection{Stage 2: Token-robust trajectory SFT}
 
Reading a single block, as in Stage~1, is not the same as reasoning over a whole
sequence of them: acting requires the model to decide which graph view to request next
and to combine the blocks it gets back. Moreover, if we simply imitate trajectories
on clean encodings, the model tends to latch onto easy surface cues in a block and
its reading collapses the moment the block changes. Stage~2 therefore teaches the
model to \emph{act}, to emit
graph-token actions and derive the answer from what they return, while keeping that
reading robust, over a corpus of
synthesized trajectories, each pairing a training anchor with a variable-length sequence of actions from $\mathcal{A}$ (Table~\ref{tab:actions}) and the
correct final answer. Training has two phases. A warm-up is ordinary supervised
fine-tuning that reproduces each trajectory's action calls and answer under a
token-weighted cross-entropy over the model's own positions $\mathcal{S}$ (the
injected blocks are inputs and are excluded),
\begin{equation}  \mathcal{L}_{\text{CE}} \;=\; \frac{1}{|\mathcal{S}|}\sum_{t\in\mathcal{S}}
  w_t\,\mathrm{CE}\!\bigl(\ell_t, y_t\bigr),
\end{equation}where $w_t=\omega$ on answer tokens and $1$ otherwise ($\omega{=}20$): upweighting
the answer tokens keeps the surface format from crowding out the decision.
Because warm-up sees only clean encodings, the reading it learns is brittle; a
consistency phase repairs this. We form two encodings of each sampled graph view, a
clean one and an augmented one that drops its edges with probability
$p_{\text{ed}}{=}0.1$ and masks graph tokens with probability $p_{\text{tm}}{=}0.2$, and
add a term that aligns their response distributions:
\begin{equation}  \mathcal{L}_2 \;=\; \mathcal{L}_{\text{CE}}^{\text{clean}}
  \;+\; \lambda_{\text{KL}}\,
  \mathrm{KL}\!\Bigl(\mathrm{sg}\bigl[p^{\text{clean}}\bigr]\,\big\|\,p^{\text{aug}}\Bigr),
\end{equation}where $p^{\text{clean}}$ and $p^{\text{aug}}$ are the softmaxed response
distributions of the two encodings, $\mathrm{sg}[\cdot]$ is the stop-gradient that makes
the clean encoding a fixed teacher, the KL runs over $\mathcal{S}$, and
$\lambda_{\text{KL}}{=}0.5$. Because the augmentations perturb exactly the graph
signal, matching the two distributions forces the model to rely on token content
that survives perturbation. Both phases update $f^{\mathrm{GNN}}$ and
$f^{\mathrm{LLM}}$ jointly.
 
\subsection{Stage 3: Graph-text-consistent IPO}
 
Stages~1 and~2 make the graph tokens readable and robust, but they do not compel
the model to \emph{use} them. The anchor's own text sits in the prompt and is often
enough to guess the label by itself, so the policy can score well while treating the
graph tokens as optional, the very shortcut we set out to avoid. Indeed,
GraphLLMs are known to lean heavily on node text, to the point that perturbing a few
words of it can flip their predictions~\cite{trustglm2025}. Supervised
fine-tuning cannot remove this shortcut: it only imitates correct trajectories and never shows
the model a case where the text and the graph point different ways, so it gives no
signal about which to trust. Stage~3 supplies that signal by learning from the
policy's own behaviour. Working from the Stage~2 policy's rollouts, it rewards
trajectories in which the graph-token evidence and the node-text evidence agree over
ones where they disagree, incentivising the policy to ground its decision in
graph-text consistency. We build preference pairs
around the \emph{consistency between a node's input text and its graph tokens}: the
preferred trajectory is one where the two agree, and the dispreferred one is
obtained by corrupting the input text so that it disagrees with the graph. For each
training anchor $v$ we run two greedy rollouts of the current policy:
\begin{itemize}[leftmargin=1.2em,itemsep=1pt,topsep=2pt]
\item a \emph{consistent} rollout $\tau^{\mathrm{c}}$ on $v$'s real node text and
real graph, where the input text attribute and the graph tokens describe the same
node and therefore agree, giving answer $A^{\mathrm{c}}$;
\item an \emph{inconsistent} rollout $\tau^{\mathrm{i}}$ that corrupts only the input
text attribute, replacing $v$'s node text with that of a randomly chosen other node
$v'$ while leaving $v$'s graph (its neighbours and their texts) unchanged, so the
text now disagrees with the graph tokens, giving answer $A^{\mathrm{i}}$.
\end{itemize}
We keep the pair only when all three conditions hold: the consistent answer is
correct ($A^{\mathrm{c}}{=}y$), the inconsistent answer is wrong
($A^{\mathrm{i}}{\neq}y$), and the two answers differ
($A^{\mathrm{c}}{\neq}A^{\mathrm{i}}$). Such a case is exactly one where making the
input text disagree with the graph flips a correct decision, so the answer had to
rest on the graph: the consistent rollout shows the graph-grounded behaviour we want
and the inconsistent one the text-reliant failure we do not. We therefore set the
preference $\tau^{+}{=}\tau^{\mathrm{c}}\succ\tau^{-}{=}\tau^{\mathrm{i}}$, which
yields on the order of $300$ pairs per run, and preferring $\tau^{+}$ trains the
policy to trust the graph over a misleading anchor text. We optimise the pairs with
Identity Preference Optimisation~\cite{ipo2024},
\begin{equation}\begin{gathered}
  \mathcal{L}_3 \;=\; \mathbb{E}_{(\tau^{+},\tau^{-})}\!\left[
  \Bigl(h(\tau^{+})-h(\tau^{-})-\tfrac{1}{2\beta}\Bigr)^{2}\right], \\[2pt]
  h(\tau)=\log\frac{f^{\mathrm{LLM}}(\tau\mid Q)}{f^{\mathrm{LLM}}_{\text{ref}}(\tau\mid Q)} ,
\end{gathered}
\end{equation}taking the Stage~2 model as $f^{\mathrm{LLM}}_{\text{ref}}$ with
$\beta{=}0.3$. As in the earlier stages, this objective updates the graph encoder
$f^{\mathrm{GNN}}$ and the language model $f^{\mathrm{LLM}}$ jointly (mpnet features
and topology frozen). Unlike preference optimisation that ranks
by human or quality judgements, our pairs are constructed to isolate
graph-dependence, so the preference directly trains graph-grounding; we use IPO over
DPO~\cite{dpo2023} because the small pair set is brittle to DPO's unbounded
log-ratio, whereas IPO's squared objective stays bounded.

\section{Experiments}

\subsection{Setup}
\label{sec:setup}

\paragraph{Datasets} We use ten text-attributed graphs across four families:
citation networks (ogbn-arxiv~\cite{ogb2020}, arXiv-2023~\cite{tape2024}, and
PubMed, Cora, CiteSeer~\cite{planetoid2016}), Amazon co-purchase and co-review graphs
(ogbn-products~\cite{ogb2020}, Amazon-Computers, Amazon-Photo~\cite{shchur2018}), a
social graph (Reddit~\cite{graphsage2017}), and a protein-interaction graph
(STRING-db~\cite{stringdb2023}).
Seven are used for in-domain training and evaluation (ogbn-arxiv, ogbn-products,
PubMed, Reddit, arXiv-2023, CiteSeer, STRING-db), and a zero-shot study transfers the
ogbn-arxiv-trained policy, without any further training, to seven targets it has never seen
(Table~\ref{tab:zeroshot}). Per-dataset node
text, label spaces, sizes, and sources are given in \textbf{Appendix~\ref{app:datasets}}.

\paragraph{Policy and encoder} Qwen2.5-3B-Instruct~\cite{qwen2024}, hidden $= 2{,}048$.
Encoder: mpnet sentence embeddings~\cite{mpnet2020,sentencebert2019} (768-d) + 2-layer GraphSAGE~\cite{graphsage2017} (hidden $512$, output
$768$) with a cross-attention pool that emits a $128$-token block per action, linearly
projected to the backbone's embedding width (our default). The 3B models train with DDP across $4\times$ RTX 5880 Ada (48~GB); the 7B
models train on H800 GPUs. Stage~3 is single-GPU with gradient accumulation. Per-stage
training corpora (construction, format, and sizes) are detailed in
\textbf{Appendix~\ref{app:corpus}}.

\paragraph{Baselines} We compare against a series of baselines: AgentGL-3B and AgentGL-7B~\cite{agentgl2026} (the RL-trained agentic
retrieval baseline), the LLaGA~\cite{llaga2024} static-graph-token baseline,
the GOFA~\cite{gofa2025} graph foundation model, and GNN references
(GraphSAGE~\cite{graphsage2017}, GCN~\cite{gcn2017}, and GAT~\cite{gat2018}).
All of these are run on the same $1{,}000$-node test samples and the same splits as our
method, so every number in Table~\ref{tab:main_xdomain} is measured under one protocol. AgentGL-7B is the strongest reported agent. We additionally re-implement seven same-backbone baselines on our exact splits (both
Qwen2.5-3B and 7B): Qwen zero-shot, Qwen chain-of-thought~\cite{cot2022}, a graph-to-text SFT model,
LLM-GNN~\cite{llmgnn2024}, and the graph-token methods TEA-GLM~\cite{teaglm2024},
GraphGPT~\cite{graphgpt2024}, and GraphTranslator~\cite{graphtranslator2024}; full
details for these are in \textbf{Appendix~\ref{app:baselines}}.

\subsection{Main results}

\begin{table*}[h]
\centering
\caption{Main results: in-domain \textbf{node classification} (\% accuracy)
and masked \textbf{link prediction} (AUC$\times100$). \textbf{Bold} indicates the
best result under the same backbone. 3-run average is reported.}
\label{tab:main_xdomain}
\scriptsize
\setlength{\tabcolsep}{3pt}
\renewcommand{\arraystretch}{0.9}
\resizebox{0.88\textwidth}{!}{%
\begin{tabular}{l ccccccc c ccccccc}
\toprule
& \multicolumn{7}{c}{\textbf{Node classification} (\% acc)} & & \multicolumn{7}{c}{\textbf{Link prediction} (AUC$\times100$)} \\
\cmidrule(lr){2-8}\cmidrule(lr){10-16}
Method & arxiv & prod & pub & reddit & a23 & cite & STR & & arxiv & prod & pub & reddit & a23 & cite & STR \\
\midrule
\multicolumn{16}{c}{\textbf{Graph Neural Network}}\\
\midrule
GraphSAGE & 69.2 & 75.9 & 88.3 & 89.8 & 66.3 & 76.5 & 68.5 & & 90.2 & 79.9 & 83.9 & 92.5 & 86.4 & 86.4 & 86.6 \\
GCN       & 68.8 & 76.1 & 86.8 & 90.5 & 66.8 & 76.5 & 62.8 & & 93.9 & 90.8 & 93.2 & 94.0 & 85.4 & 91.0 & 91.8 \\
GAT       & 66.9 & 75.8 & 84.6 & 89.5 & 67.7 & 78.1 & 62.4 & & 87.0 & 88.2 & 84.7 & 95.4 & 87.0 & 89.5 & 89.5 \\
\midrule
\multicolumn{16}{c}{\textbf{Backbone: Qwen2.5-3B-Instruct}}\\
\midrule
LLaGA   & 66.9 & 71.7 & 86.5 & 88.3 & 62.9 & \textbf{75.9} & 50.6 & & 82.2 & 87.8 & 88.2 & 88.7 & \textbf{94.7} & 76.3 & 79.7 \\
AgentGL & 63.9 & 64.4 & 89.1 & 76.6 & 29.7 & 71.5 & 39.8 & & 71.0 & 61.3 & 77.9 & 63.1 & 72.3 & 71.1 & 71.0 \\
GOFA    & 63.9 & 65.1 & 77.9 & 83.5 & 41.2 & 63.0 & 28.4 & & 79.7 & 80.7 & 67.0 & 73.8 & 71.6 & 71.4 & 56.3 \\
Qwen (zero-shot) & 33.0 & 50.5 & 74.6 & 45.5 & 28.8 & 58.8 & 36.3 & & 64.8 & 72.8 & 85.6 & 57.5 & 73.3 & 72.7 & 75.2 \\
Chain-of-Thought       & 60.4 & 49.3 & 82.3 & 46.2 & 56.4 & 61.4 & 38.7 & & 65.7 & 75.8 & 88.1 & 65.8 & 74.5 & 72.9 & 76.2 \\
Graph2Text-SFT   & 69.3 & 75.0 & 92.9 & 89.4 & 73.7 & 72.3 & 68.6 & & 83.7 & 90.5 & 92.6 & 94.2 & 90.8 & 88.7 & 85.1 \\
LLM-GNN          & 48.9 & 53.2 & 61.3 & 50.3 & 36.0 & 63.2 & 33.2 & & 53.1 & 53.4 & 53.1 & 50.9 & 50.8 & 52.9 & 48.5 \\
TEA-GLM      & 70.4 & 70.3 & 89.9 & 53.5 & 63.0 & 50.3 & 18.5 & & 85.7 & 94.7 & 85.8 & 93.0 & 86.6 & 87.9 & 86.6 \\
GraphGPT      & 70.3 & 74.0 & 93.0 & 68.3 & 69.9 & 52.8 & 42.3 & & 85.6 & 93.7 & 87.9 & 93.3 & 86.3 & 89.4 & 86.1 \\
GraphTranslator & 71.5 & 73.6 & 86.7 & 77.5 & 69.6 & 74.5 & 35.7 & & 84.6 & 93.4 & 92.2 & 95.3 & 88.8 & 87.6 & 87.4 \\
\midrule
\textbf{Ours} & \textbf{73.0} & \textbf{76.8} & \textbf{93.3} & \textbf{96.9} & \textbf{77.5} & 73.8 & \textbf{69.1} & & \textbf{97.7} & \textbf{96.7} & \textbf{96.4} & \textbf{97.8} & 92.0 & \textbf{96.8} & \textbf{88.5} \\
\midrule
\multicolumn{16}{c}{\textbf{Backbone: Qwen2.5-7B-Instruct}}\\
\midrule
LLaGA   & 68.2 & 72.6 & 90.5 & 92.4 & 65.4 & 76.3 & 51.5 & & 82.3 & 87.3 & 88.9 & 88.7 & 96.5 & 75.6 & 80.3 \\
AgentGL & 69.5 & 65.3 & 90.8 & 88.3 & 58.3 & 73.8 & 43.1 & & 74.2 & 66.6 & 78.4 & 62.0 & 73.8 & 70.8 & 71.4 \\
GOFA    & 67.2 & 66.8 & 84.9 & 85.7 & 59.2 & 72.4 & 34.3 & & 82.9 & 80.8 & 69.3 & 75.2 & 72.6 & 73.5 & 58.3 \\
Qwen (zero-shot) & 66.5 & 47.3 & 90.1 & 47.1 & 65.8 & 59.6 & 43.3 & & 92.8 & 92.3 & 94.7 & 76.1 & 96.6 & 94.9 & 85.4 \\
Chain-of-Thought       & 67.7 & 50.5 & 91.2 & 48.2 & 59.8 & 73.3 & 43.5 & & 93.5 & 92.8 & 94.2 & 78.7 & 96.4 & 95.1 & 86.6 \\
Graph2Text-SFT   & 70.8 & 75.8 & 93.6 & 93.7 & 75.5 & 75.9 & 69.5 & & 94.5 & 96.9 & 95.7 & 95.7 & 96.6 & 95.7 & 88.0 \\
LLM-GNN          & 65.8 & 43.3 & 77.6 & 49.9 & 55.2 & 58.8 & 33.9 & & 57.0 & 52.3 & 50.8 & 52.0 & 51.0 & 54.1 & 50.5 \\
InstructGLM      & 69.8 & 66.9 & 88.1 & 49.5 & 62.9 & 72.4 & 21.5 & & 71.9 & 73.1 & 96.2 & 57.5 & 90.6 & 83.3 & 72.7 \\
TEA-GLM      & 71.5 & 74.0 & 82.1 & 77.6 & 59.5 & 55.2 & 48.5 & & 87.7 & 95.5 & 87.5 & 93.6 & 87.1 & 88.8 & 86.4 \\
GraphGPT      & 67.0 & 72.9 & 93.4 & 93.3 & 67.0 & 73.8 & 50.6 & & 64.7 & 94.2 & 60.0 & 76.4 & 78.1 & 63.7 & 77.5 \\
GraphTranslator & 74.4 & 77.8 & 90.0 & 79.8 & 70.1 & 45.3 & 39.2 & & 90.1 & 97.4 & 91.8 & 98.0 & 89.9 & 93.6 & 88.9 \\
\midrule
\textbf{Ours} & \textbf{75.6} & \textbf{80.1} & \textbf{95.6} & \textbf{98.2} & \textbf{77.6} & \textbf{78.8} & \textbf{71.3} & & \textbf{97.9} & \textbf{98.8} & \textbf{96.7} & \textbf{98.1} & \textbf{96.9} & \textbf{96.7} & \textbf{89.2} \\
\bottomrule
\end{tabular}%
}
\end{table*}

Our 3B model sets a new state of the art among comparably sized graph LLMs
(Table~\ref{tab:main_xdomain}). On node classification it leads on six of the seven
domains, reaching $73.0\%$ on ogbn-arxiv and beating the comparably sized
AgentGL-3B baseline ($63.9\%$) by $9.1$ points; we show later
(Figure~\ref{fig:error-density-pp}) that this gain concentrates on the challenging
nodes, where the anchor text alone is ambiguous, rather than on the easy ones. On
masked link prediction it wins six of the seven domains, trailing only LLaGA on
arXiv-2023. In fact LLaGA and GraphTranslator are the only trained graph-LLM baselines that beat
ours on any domain (LLaGA on arXiv-2023 LP and CiteSeer NC, GraphTranslator on
CiteSeer NC), while the retrieval-based AgentGL and
frozen-feature GOFA sit well below. A plain GNN stays ahead of ours only on the two
most structure-dominated cases, the homophilous CiteSeer (NC) and the
STRING-db (LP) graphs.

The advantage is consistent at larger scale. With a Qwen2.5-7B backbone our model tops
every node-classification domain (e.g.\ $75.6\%$ on ogbn-arxiv and $80.1\%$ on
ogbn-products) and every link-prediction domain among the same-backbone baselines, so
it consistently outperforms every same-backbone baseline across both tasks and the
approach scales with backbone size.

\subsection{Zero-shot transfer}

\begin{table}[h]
\centering
\caption{Zero-shot accuracy: every method trained only on
ogbn-arxiv and evaluated on target with no fine-tuning. }
\label{tab:zeroshot}
\small
\setlength{\tabcolsep}{3pt}
\resizebox{\columnwidth}{!}{%
\begin{tabular}{lccccccc}
\toprule
Method & Cora & Cite. & Pub. & prod. & a-comp & a-photo & STR. \\
\midrule
LLaGA        & 4.6  & 19.0 & 49.0 & 1.4  & 4.2  & 0.4  & 0.7  \\
AgentGL      & 42.4 & 40.4 & 39.8 & 36.6 & 41.2 & 40.7 & 21.0 \\
GOFA         & 26.3 & 16.9 & 41.0 & 11.7 & 10.3 & 5.5  & 1.4  \\
TEA-GLM      & 49.2 & 51.9 & 56.3 & 35.7 & 25.4 & 43.3 & 20.0 \\
GraphGPT     & 33.0 & 40.0 & 20.5 & 32.5 & 36.6 & 46.6 & 7.7  \\
GraphTrans.  & 34.8 & 51.3 & 53.0 & 33.2 & 38.9 & \textbf{48.4} & 18.4 \\
\textbf{Ours} & \textbf{68.9} & \textbf{54.4} & \textbf{82.0} & \textbf{55.8} & \textbf{53.3} & 43.7 & \textbf{37.4} \\
\bottomrule
\end{tabular}%
}
\end{table}

We use the arxiv checkpoint without any further training to validate zero-shot performance. Transfer quality (Table~\ref{tab:zeroshot}) tracks the results: PubMed's three
broad medical categories admit reliable text classification ($82.0\%$),
while the 21-class STRING-db is hardest ($37.4\%$). Ours leads on six of the seven
transfer targets, trailing only Amazon-Photo (won by GraphTranslator, $48.4$ vs
$43.7$), and by large margins over LLaGA, AgentGL, and GOFA.

\subsection{Ablations}

\begin{table*}[h]
\centering
\caption{Component ablation (Qwen2.5-3B). N/A marks an action outside a task's action space.}
\label{tab:ablation_xdomain}
\footnotesize
\setlength{\tabcolsep}{3pt}
\resizebox{0.82\textwidth}{!}{%
\begin{tabular}{l ccccccc c ccccccc}
\toprule
& \multicolumn{7}{c}{\textbf{Node classification} (\% acc)} & & \multicolumn{7}{c}{\textbf{Link prediction} (AUC$\times100$)} \\
\cmidrule(lr){2-8}\cmidrule(lr){10-16}
Configuration & arxiv & prod & pub & reddit & a23 & cite & STR & & arxiv & prod & pub & reddit & a23 & cite & STR \\
\midrule
\textbf{Full model} & \textbf{73.0} & \textbf{76.8} & \textbf{93.3} & \textbf{96.9} & \textbf{77.5} & \textbf{73.8} & \textbf{69.1} & & \textbf{97.7} & \textbf{96.7} & \textbf{96.4} & \textbf{97.8} & \textbf{92.0} & \textbf{96.8} & \textbf{88.5} \\
\midrule
\multicolumn{16}{@{}l}{\emph{Training stages}} \\
\quad w/o Stage~2 consistency (CRJT)            & 70.5 & 76.4 & 93.0 & 93.2 & 75.4 & 72.4 & 66.4 & & 87.4 & 92.4 & 85.2 & 97.4 & 91.7 & 88.5 & 87.5 \\
\quad w/o Stage~3 (IPO) & 71.9 & 75.9 & 92.7 & 95.4 & 76.3 & 73.2 & 68.6 & & 89.5 & 90.3 & 88.1 & 97.2 & 91.5 & 89.0 & 88.1 \\
\addlinespace[2pt]
\multicolumn{16}{@{}l}{\emph{Action space}} \\
\quad w/o retrieval token             & 72.8 & 76.2 & 92.8 & 96.0 & 76.7 & 72.9 & 69.0 & & 97.2 & 96.1 & 96.0 & 97.3 & 91.2 & 96.4 & 88.1 \\
\quad w/o cluster token               & 72.7 & 75.5 & 92.7 & 96.4 & 77.1 & 72.8 & 68.6 & & 97.2 & 96.0 & 96.3 & 97.2 & 91.1 & 96.2 & 86.9 \\
\quad w/o node token                  & 71.8 & 75.9 & 92.1 & 96.5 & 76.2 & 72.8 & 68.5 & & 95.6 & 86.7 & 83.0 & 97.5 & 90.1 & 95.0 & 86.7 \\
\quad w/o neighbour tokens (1/2/3-hop)  & 70.2 & 76.2 & 92.4 & 96.3 & 77.1 & 71.8 & 68.8 & & 95.4 & 94.2 & 94.8 & 95.7 & 90.5 & 96.0 & 80.1 \\
\quad w/o cosine similarity        & N/A & N/A & N/A & N/A & N/A & N/A & N/A & & 97.0 & 64.1 & 84.6 & 97.0 & 82.6 & 95.5 & 85.8 \\
\bottomrule
\end{tabular}%
}
\end{table*}

\paragraph{Stage components} Table~\ref{tab:ablation_xdomain} isolates the
Stage~2 consistency term (CRJT) from full model; averaged over domains, it
is the largest of the training components ($1.9$ points on node classification and $5.1$ on
link prediction). CRJT pushes the LM to be invariant to two augmentations (edge-drop and
token-mask) that the encoder cannot itself control, forcing the projection head and
the LM's attention into the graph-token block to extract scope-invariant signal.
Without it the LM treats graph tokens as a weak prior; with it the model calls
\verb|cluster_token| and \verb|three_hop_token| far more often at greedy decoding,
which translates directly into accuracy. Every domain improves with CRJT, with the
largest gains on link prediction
(Table~\ref{tab:ablation_xdomain}).

\paragraph{Action contributions} The lower block of Table~\ref{tab:ablation_xdomain}
removes one action at a time. For node classification the neighbour tokens (1/2/3-hop)
matter most: dropping them causes the largest accuracy fall, confirming that the
topological signal, not the anchor alone, drives the label. The single-node, retrieval, and
cluster tokens each contribute smaller amounts; on the $1{,}000$-node samples several of these
node-classification differences are within measurement noise, and we treat them as indicative
rather than individually significant. For link prediction
the picture shifts, as a link decision rests on comparing the two endpoints: the
cosine-similarity action becomes decisive and the single-node token is next most
important, both collapsing AUC sharply when ablated (the cosine action costs $32.6$ AUC on
ogbn-products, the node token $13.4$ on PubMed).
Every action contributes on at least one task, which supports keeping the full action
space rather than a fixed subset.

\subsection{Is graph token better for reasoning?}

Our central claim is that the \emph{medium}, encoding each reasoning step as a
graph token rather than as text, is what drives the gains, not the agentic loop
alone. To test this directly, we build a text-medium ablation that shares our exact
agentic framework: the same action space (Table~\ref{tab:actions}), the same
step-by-step rollout, and the same Stage~2--3 training recipe, changing only what each
action returns. Stage~1 is skipped for the text medium, since there is no graph-token
block to ground. The verbaliser renders the same graph view the encoder would have pooled
and is subject to the same $2048$-token context limit used throughout; the serialised
views fit within that budget, so the text medium is not penalised by truncation.
Instead of an AGT block $\mathbf{Z}^{\mathrm{AGT}}_t$, the encoder is
replaced by a verbaliser that serialises the named graph view $G^{V_t}$ into
natural-language text spliced into the context. Any gap between the two rows in
Table~\ref{tab:medium} is attributable to the medium alone. In domain the two media stay close (81.0 versus 77.8 on average, a 3.2-point gap),
where the anchor text alone often settles the label, but they diverge sharply under transfer.
The graph-token medium retains $57.4\%$ zero-shot accuracy against $28.1\%$ for the
verbalised-text medium, a 29-point gap on the same targets. The verbaliser preserves
the same neighbourhood the encoder sees, so the difference is not about which graph view
is read but about how it is represented: serialising to text discards the compact
structural signal the encoder carries across datasets, which is precisely where
out-of-domain accuracy depends on it. Because the text medium keeps our trajectory SFT
and rollout unchanged and only forgoes the pooler it has no use for, this gap is an estimate of what the medium contributes.

\begin{table}[h]
\centering
\small
\caption{Isolating the graph-token medium.
Our agentic framework is run with each action returning either verbalised \emph{text}
(\textsc{Ours-Text}) or a \emph{graph token} (\textsc{Ours-Token}), with identical actions; $\Delta$ is the graph-token gain over text.}
\label{tab:medium}
\begin{tabular}{lcccc}
\toprule
 & AgentGL & Ours-Text & Ours-Token & $\Delta$ \\
\midrule
\multicolumn{5}{@{}l}{\emph{In-domain}} \\
arxiv      & 63.9 & 71.9 & \textbf{73.0} & $+1.1$ \\
products   & 64.4 & 71.3 & \textbf{76.8} & $+5.5$ \\
pubmed     & 89.1 & 90.1 & \textbf{93.3} & $+3.2$ \\
average    & 72.5 & 77.8 & \textbf{81.0} & $+3.2$ \\
\midrule
\multicolumn{5}{@{}l}{\emph{Zero-shot transfer}} \\
pubmed     & 39.8 & 32.6 & \textbf{82.0} & $+49.4$ \\
products   & 36.6 & 43.7 & \textbf{55.8} & $+12.1$ \\
citeseer   & 40.4 & 31.8 & \textbf{54.4} & $+22.6$ \\
STRING-db  & 21.0 & 4.3  & \textbf{37.4} & $+33.1$ \\
average    & 34.5 & 28.1 & \textbf{57.4} & $+29.3$ \\
\bottomrule
\end{tabular}
\end{table}

\begin{figure}[h]
\centering
\includegraphics[width=\columnwidth]{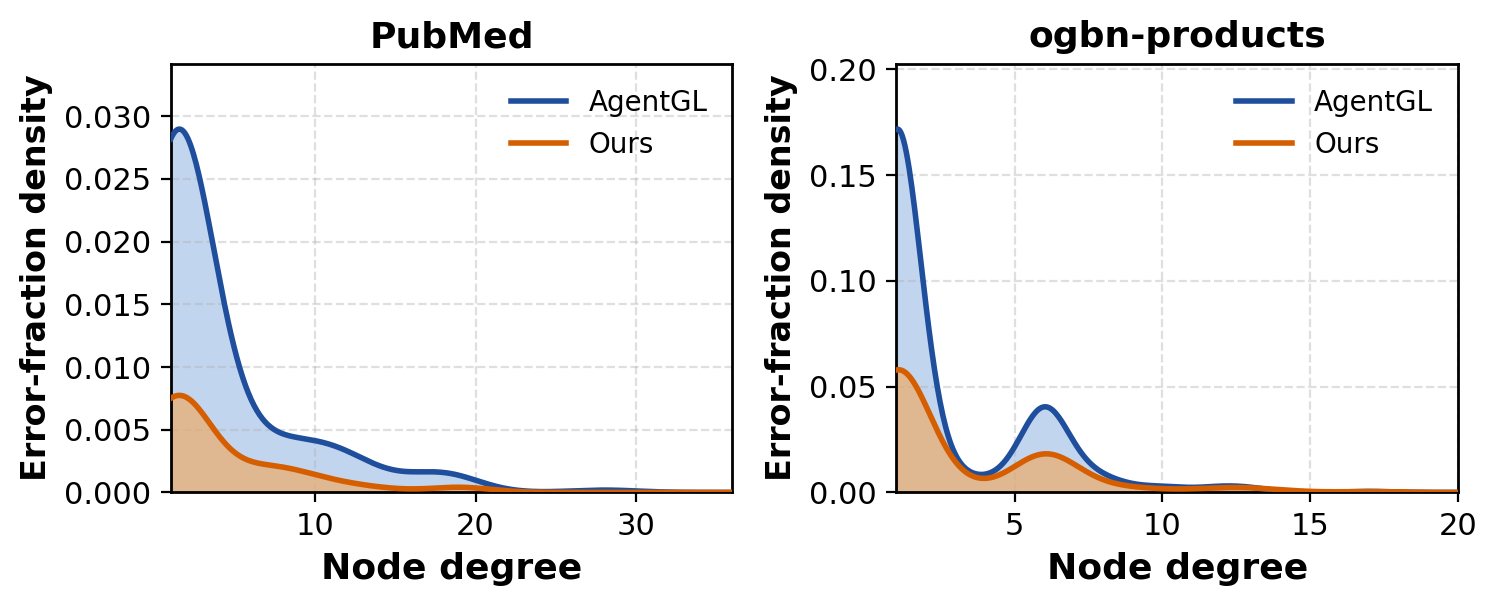}
\caption{Error-fraction density over node degree: \textsc{Ours-Token} (orange) versus the
AgentGL text agent (blue). A lower curve
means fewer and better-placed errors. The full analysis is in
Figure~\ref{fig:error-density}.}
\label{fig:error-density-pp}
\end{figure}

\paragraph{Where the errors reduce.} The medium's advantage is spread across nodes, not
confined to a few. Figure~\ref{fig:error-density-pp} plots the density of misclassified
nodes over node degree for \textsc{Ours-Token} against the AgentGL text agent on
\textsc{PubMed} and \textsc{ogbn-products}, each curve scaled to its total error mass. Our
error mass on PubMed is about three-fifths of the text agent's and on ogbn-products about
two-thirds, removing roughly $38\%$ and $35\%$ of its errors. Both methods err most on low-degree nodes,
where the neighbourhood carries the least signal, yet our curve stays below the text agent
across the entire degree range, not only on well-connected hubs. The graph-token
medium thus lowers error mass broadly, which is consistent with the
transfer gap in Table~\ref{tab:medium}. \textbf{Appendix~\ref{app:error}} extends this analysis (Figure~\ref{fig:error-density}).

\subsection{Pareto frontier} 

\begin{figure}[t]
\centering
\includegraphics[width=0.82\columnwidth]{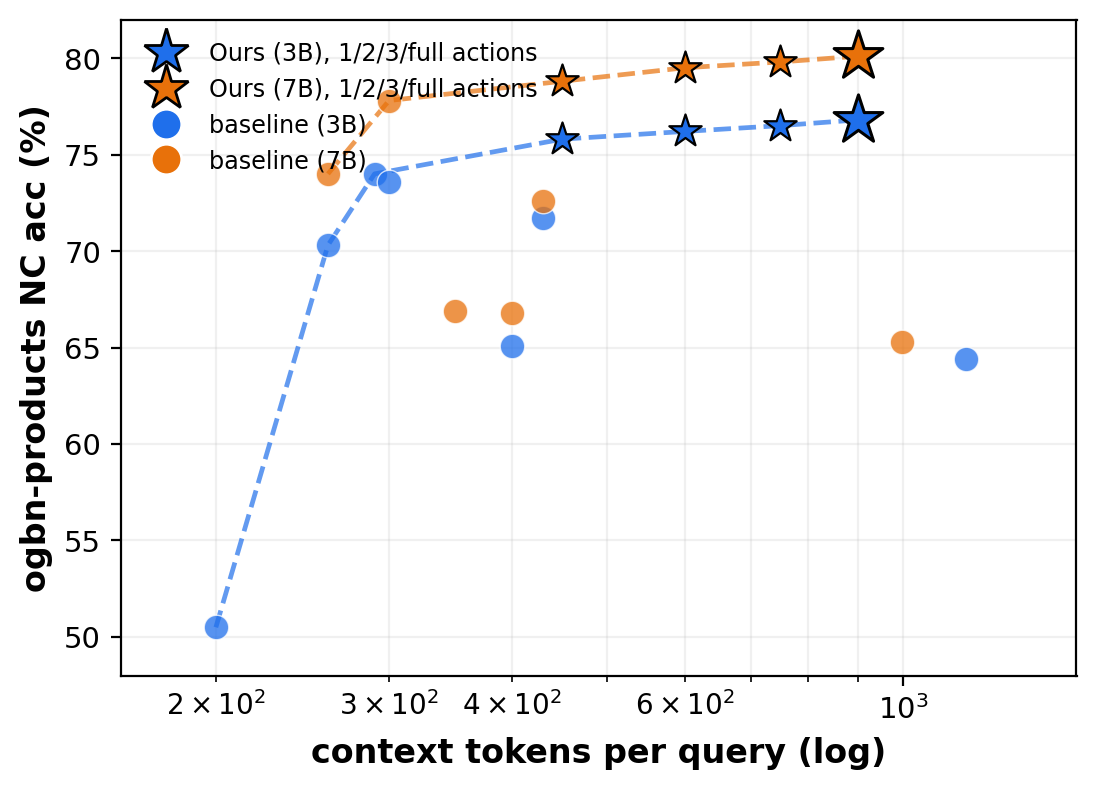}
\caption{Accuracy versus inference tokens on \textsc{ogbn-products}. Blue $3$B, orange $7$B; Star: ours, circle: baseline, and the dashed line is the Pareto frontier for each size. More results
are in \textbf{Appendix~\ref{app:pareto-perdataset}}.}
\label{fig:pareto-products}
\end{figure}

Figure~\ref{fig:pareto-products} places the methods on the
accuracy versus token consumption for \textsc{ogbn-products}, with a dashed line tracing
the Pareto frontier for each backbone size; we zoom into the informative larger-context region.
Our $3$B and $7$B models occupy the entire high-accuracy end of their frontier while holding a
bounded budget (near $900$ tokens for 4 actions), because each action contributes a fixed-size graph-token block
rather than growing with the retrieved text. To make the trade-off explicit we also plot
checkpoints constrained to emit at most one, two, or three actions: the
one-action variant already lands above every baseline, and
accuracy climbs by roughly a point as the budget grows to the full trajectory before saturating,
so a caller can trade a few tokens for a small accuracy change without leaving the frontier. The
retrieval-based AgentGL spends the most context yet trails in accuracy, and the single-pass
graph-token baselines are cheaper but several points lower. The full per-dataset analysis is in \textbf{Appendix~\ref{app:pareto-perdataset}}
(Figures~\ref{fig:pareto-perdataset} and~\ref{fig:pareto-perdataset-b}).

\subsection{Understanding the learned policy}

We inspect the trained policy on ogbn-arxiv, where it issues on average 4 actions per
trajectory. Three findings emerge. First, the policy is adaptive: trajectory length is
stable within ogbn-arxiv but ranges from a single action to the full ladder across
datasets, lengthening with how much the node text leaves unresolved, so the model reaches
for graph tokens only when the text falls short. Second, the graph view it chooses to encode
carries the discriminative signal the text alone leaves ambiguous. Third, the injected
tokens are legible: passed through a Stage~1 reader, a token decodes to the research area of
its graph view, and that topical signal is what fixes the label.
The activity-recognition paper ``gimme signals'', whose abstract reads like signal
processing, is a representative case: its tokens decode to computer-vision content and move
the prediction to \texttt{cs.CV}. More analysis results are in \textbf{Appendix~\ref{app:case-studies}}.

\subsection{Further analysis} 

More studies are deferred to the appendix for space.
\textbf{Appendix~\ref{app:time}} analyses training cost, placing accuracy against total training time and
breaking our three-stage pipeline down stage by stage. \textbf{Appendix~\ref{app:sensitivity}} reports
sensitivity checks on hyperparameters.

\section{Conclusion}

We introduced \emph{Agentic Graph Token (AGT) reasoning}, which turns graph token
from a one-shot preprocessing step into part of reasoning: at each step the model names
a graph view, an encoder materialises it into a fixed-size graph-token block on demand,
and the block is spliced back into the context, so the tokens it reads are chosen as it
reasons. A three-stage pipeline teaches the model to read graph tokens, keeps that reading
grounded through a consistency regulariser, and rewards trajectories whose graph and text
evidence agree. Our model outperforms comparably sized graph-token and
agentic baselines on graph benchmarks and transfers zero-shot to
unrelated domains, with our analyses attributing the gains to the graph-token medium
rather than the agentic loop alone. Taken together, the results argue for treating the
graph tokens as an instrument the model calls during reasoning rather than a fixed
pattern, pushing graph analysis to an effective agentic paradigm.

\bibliographystyle{ACM-Reference-Format}
\bibliography{main}

\appendix

\section{Limitations and Future Work}
\label{app:limitations}

The node text features come from a frozen mpnet encoder throughout; a stronger
pretrained feature extractor would propagate directly through our jointly trained
graph encoder. We did not
explore joint NC $+$ LP training extensively in this work; preliminary
results suggest that node-classification and link-prediction actions need separate
namespaces before the model emits them coherently, and we leave a full joint study to
future work. The IPO stage is currently
fed only with text-shuffle preference pairs; combining text and
graph perturbations into a single pair set is a natural next step.

\section{Dataset descriptions}
\label{app:datasets}

We evaluate on the ten text-attributed graphs listed in Section~\ref{sec:setup};
each node carries raw text, from which the MPNet features are computed. Following
AgentGL~\cite{agentgl2026}, node classification samples $3{,}000$ training nodes per
dataset from the training split and evaluates on a held-out sample of $1{,}000$ test
nodes; link prediction trains on $1{,}500$ positive and $1{,}500$ negative
node pairs per dataset and tests on $500$ positive and $500$ negative pairs. We detail
each graph below.

\paragraph{Citation graphs} \textbf{ogbn-arxiv} is a citation network of arXiv CS
papers labelled with $40$ subject areas, with the title and abstract as node text; it
is our primary in-domain benchmark and the source domain for all zero-shot transfer.
\textbf{arXiv-2023} is a same-label-space ($40$-class) citation graph of papers
published in 2023, used to probe temporal distribution shift. \textbf{PubMed}
($3$ diabetes-related classes), \textbf{Cora} ($7$ machine-learning topics), and
\textbf{CiteSeer} ($6$ computer-science topics) are the standard small citation
benchmarks, again with titles/abstracts as node text.

\paragraph{Co-purchase / co-review graphs} \textbf{ogbn-products} is an Amazon
co-purchase graph with product descriptions as node text and $42$ product categories.
\textbf{Amazon-Computers} ($10$ classes) and \textbf{Amazon-Photo} ($8$ classes) are
smaller co-purchase graphs from the same family, used as zero-shot targets.

\paragraph{Social graph} \textbf{Reddit} is a post-interaction graph with post text as
node attributes and community labels.

\paragraph{Biological graph} \textbf{STRING-db} is a human protein-protein
interaction network built from the STRING database~\cite{stringdb2023}; node text describes each
protein and the $21$ labels are COG protein-function classes. It stresses methods on a
largely non-text-centric domain where the label depends heavily on graph structure.

\section{Baseline implementation details}
\label{app:baselines}

Every same-backbone baseline uses the identical Qwen2.5-3B-Instruct backbone, the
identical per-domain splits (the same train/test node splits for node
classification and the same candidate pairs for link prediction), and the same
MPNet node text-features as our method; they
differ only in mechanism. All are re-implemented faithfully to each method's
original design. We verified that no test node or candidate pair appears in any training
corpus for any domain, including the alignment corpora.

\paragraph{Qwen (zero-shot)} The frozen backbone is prompted with the target node's
title$+$abstract and the candidate-category list and asked to emit a single category.
No graph structure and no training.

\paragraph{Qwen (CoT)} As zero-shot, but the model is asked to reason step by step before
committing to a category (it names the core topic and method, shortlists the likely classes,
and eliminates the rest). Text only, no graph, no training.

\paragraph{Graph2Text-SFT} A supervised baseline that verbalises the anchor node together
with its one-hop neighbours as text and fine-tunes the Qwen backbone to predict the label.
It uses the graph structure but as serialised text rather than tokens.

\paragraph{LLM-GNN} Following the LLM-as-annotator recipe, Qwen labels a pool of
nodes from text alone; a GraphSAGE classifier is then trained on these pseudo-labels
over the graph and evaluated on the held-out test set.

\paragraph{Graph-token baselines (TEA-GLM, GraphGPT, GraphTranslator)} These share
our token interface: a node's $k$-hop neighbourhood is pooled by a GraphSAGE encoder
over MPNet features and projected to a block of $128$ graph-token embeddings that are
spliced into the prompt in place of a placeholder span. They differ in \emph{what is
trained}:
\begin{itemize}
\item \textbf{TEA-GLM} keeps the LLM frozen and trains only the graph
encoder/projector, on a single-token instruction corpus, one $1$-hop graph token
per training node plus a direct classification instruction.
\item \textbf{GraphGPT} uses the same single-token corpus but \emph{tunes} the LLM
jointly with the encoder (learning rate $10^{-5}$); it is the tuned counterpart of
frozen TEA-GLM.
\item \textbf{GraphTranslator} freezes the LLM and trains a two-stage translator:
Stage~A aligns graph tokens to an LLM-generated description of the node and its
neighbours, following the method's Producer stage, and Stage~B fine-tunes the translator
on the downstream task.
\end{itemize}

\paragraph{Common training setup} The encoder/projector are trained with graph
learning rate $5\times10^{-4}$, one epoch, gradient accumulation $4$, maximum
sequence length $2048$, and an answer-token loss weight of $20$. Node-classification
corpora are capped at $20$k training samples and link-prediction corpora at $6$k,
drawn only from the training split. All baselines run on the same hardware as our
method.

\paragraph{Link prediction} For a pair $(u,v)$, each endpoint's $1$-hop subgraph is
encoded \emph{with the target edge masked from both endpoints}, identical to our
method. The two token blocks are spliced into a yes/no edge-existence prompt and we
score $\mathrm{logit}(\texttt{yes})-\mathrm{logit}(\texttt{no})$, reporting AUC over
the balanced $1{,}000$-pair test set. TEA-GLM freezes the LLM, GraphGPT tunes it
(learning rate $10^{-5}$), and GraphTranslator uses its two-stage translator; all
train on AgentGL's canonical $1{,}500$ positive $+$ $1{,}500$ negative training pairs,
the same pairs our Stages~2--3 consume.

\section{Per-stage training corpus}
\label{app:corpus}

Our three training stages consume distinct corpora, all built from the same
per-domain training split; no test node appears in any stage. Node text-features are
MPNet embeddings of title$+$abstract, and graph tokens are produced on the fly by the
GraphSAGE encoder for whichever scope an action requests. We describe each stage's
corpus construction, format, and size below (ogbn-arxiv figures; other domains follow
the same recipe).

\paragraph{Stage 1: alignment (learning to read graph tokens)} Purpose: ground
every kind of graph token so the LM can decode node content and local structure from
the token block alone. For each training node we sample one action scope ($1$-, $2$-,
or $3$-hop subgraph pooling, top-$K$ MPNet retrieval, or cosine-similarity), encode it
to a $128$-token block, and set the target to (i)~reconstruct the anchor's
title$+$abstract and/or (ii)~answer a masked link-prediction query, from the tokens
alone. Covering all scopes ensures every token type is supervised. Corpus:
$\approx\!40$k examples, one epoch.

\paragraph{Stage 2: token-robust trajectory SFT} Purpose: teach free-choice
agentic reasoning. Each example is a multi-turn trajectory: the prompt carries the
node's title$+$abstract and the task; the assistant then emits a sequence of action
calls (\texttt{node\_token}, \texttt{one\_hop\_token}, \texttt{two\_hop\_token},
\texttt{three\_hop\_token}, \texttt{retrieval\_token}, \texttt{cluster\_token}; plus
\texttt{cosine\_sim} for link prediction), each returning a graph-token block, and
finally the answer. The loss is masked to the assistant's action and answer spans. Corpus: $\approx\!9$k trajectories, one
epoch, maximum length $2048$. To force the policy to \emph{use} the tokens rather than
lean on the prompt text, training uses consistency-regularized joint training (CRJT):
graph tokens are perturbed (token-mask $0.2$, edge-drop $0.1$) and a KL term
(weight $0.5$) aligns the clean and perturbed predictions.

\paragraph{Stage 3: rollout IPO} Purpose: refine the Stage-2 policy with
preference optimisation on its own rollouts. For each training node we run two greedy
rollouts of the Stage-2 policy: a consistent one on the real node text and graph, and
an inconsistent one whose node text is shuffled so that it disagrees with the graph. We
keep a pair only when the consistent rollout is correct and the shuffled one is wrong,
and prefer the consistent trajectory, so the objective rewards agreement between the
graph and text evidence. Preferences are optimised with IPO ($\beta=0.3$, learning rate
$3\times10^{-7}$, gradient accumulation $4$). No new graph data is introduced, the stage
reweights trajectories the model already produces.

\paragraph{Link prediction} LP shares Stage~1 and re-runs Stages~2--3 on LP-specific
corpora built from AgentGL's canonical training pairs ($1{,}500$ positive $+$ $1{,}500$
negative per domain): each example encodes both endpoints' $1$-hop subgraphs with the
target edge masked from both, and supervises a \texttt{yes}/\texttt{no} edge decision.

\section{Error structure analysis}
\label{app:error}

\begin{figure*}[t]
\centering
\includegraphics[width=0.9\textwidth]{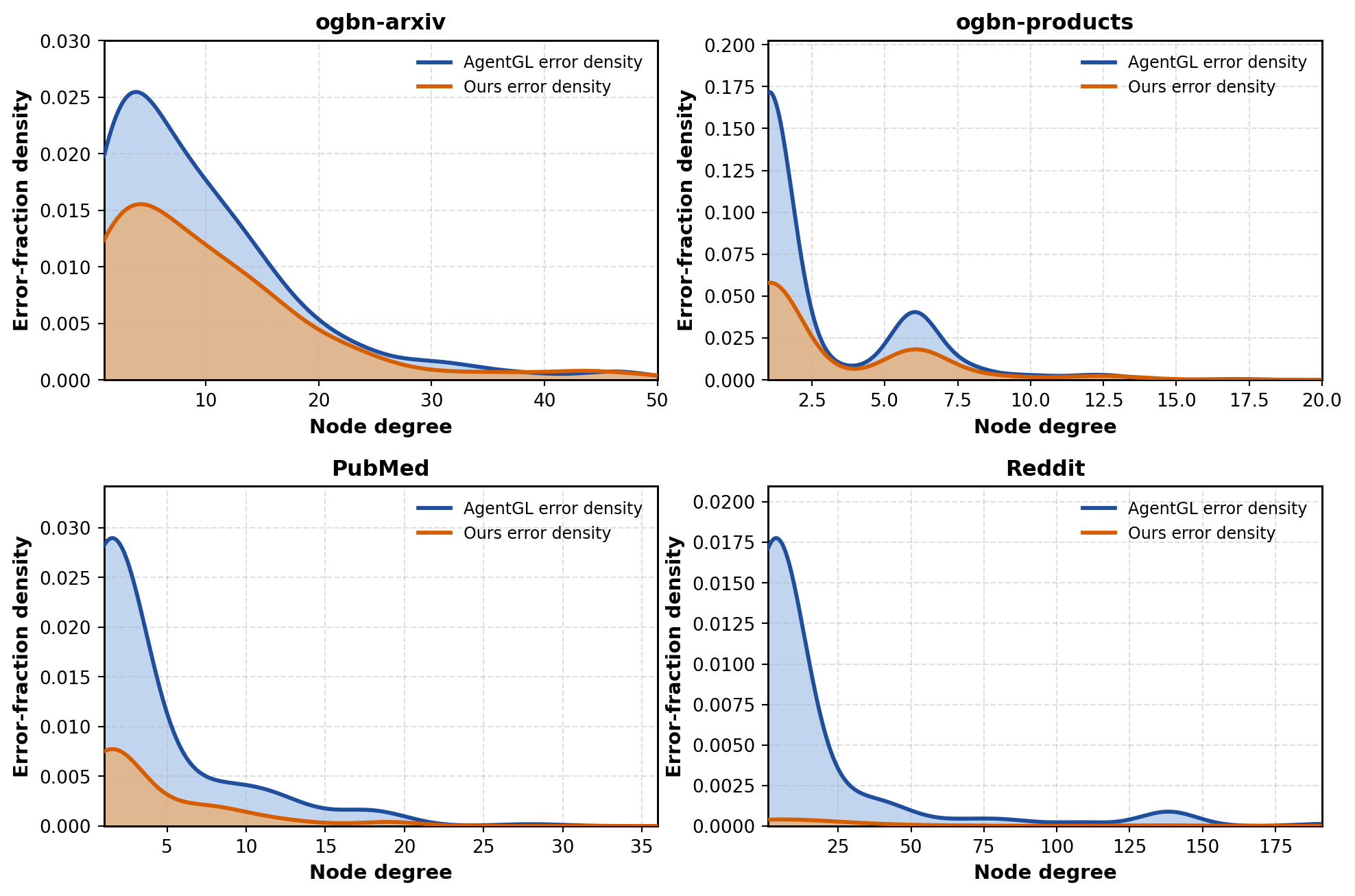}
\caption{Error-fraction density over node degree, one panel per dataset, using the same
conventions as Figure~\ref{fig:error-density-pp}. Our density lies below AgentGL's on all
four datasets.}
\label{fig:error-density}
\end{figure*}

To understand \emph{where} our agentic graph-token policy helps, we compare its
per-node error structure against AgentGL, the strongest agentic text-retrieval
baseline, on the shared $1{,}000$-node test samples of four datasets: ogbn-arxiv,
ogbn-products, PubMed, and Reddit. Both methods answer the same nodes, so the
comparison is paired. Figure~\ref{fig:error-density} plots the error-fraction density
over node degree, one panel per dataset: a Gaussian kernel density estimate over the
degree of the misclassified nodes, with each curve scaled so its area equals that
model's error mass. A lower curve therefore means fewer errors, and its shape shows on
which nodes they fall.

On every dataset our error mass sits below AgentGL's, and on the text-rich graphs the
gap is wide. The errors of both methods concentrate on lower-degree nodes, where the
graph signal is scarcest, but our density stays below AgentGL's across the whole degree
range, not merely at the well-connected end. Choosing a graph view to encode thus
leaves fewer residual errors than retrieving text, and the improvement is spread across
the degree spectrum instead of resting on a few easy hubs.

\section{Inference cost}
\label{app:pareto-perdataset}

\begin{figure*}[t]
\centering
\includegraphics[width=\textwidth]{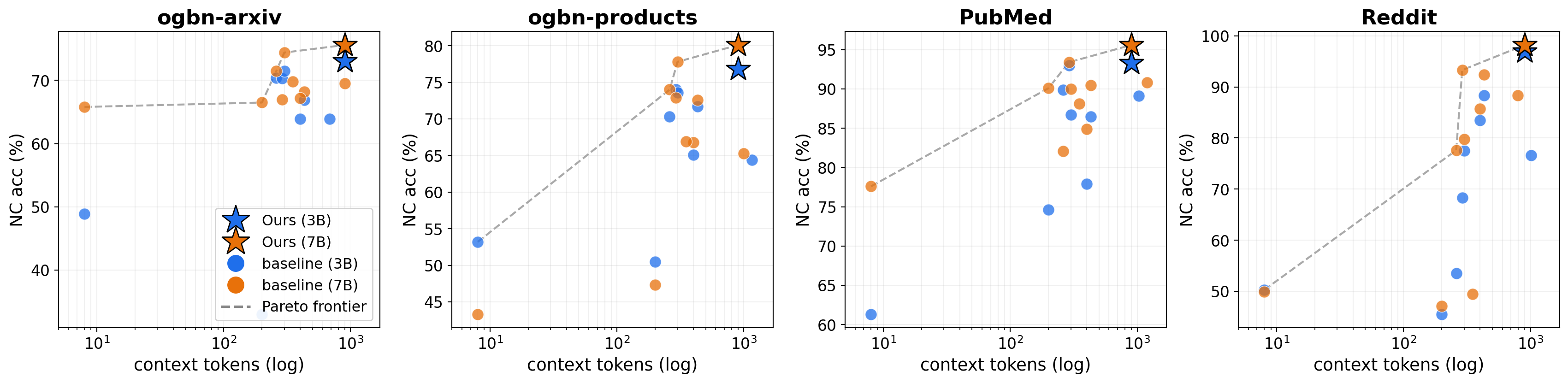}
\caption{Accuracy against inference context (part 1 of 2): ogbn-arxiv, ogbn-products,
PubMed, Reddit. Colour encodes backbone size (blue = 3B, orange = 7B) and marker encodes
method (star = ours, circle = baseline); the horizontal axis is tokens processed per query
on a log scale. AgentGL's context is measured per dataset; the other methods use a
method-representative value measured on ogbn-arxiv. Both our variants sit on the
accuracy-maximal corner of the frontier at a bounded context near $900$ tokens.}
\label{fig:pareto-perdataset}
\end{figure*}

\begin{figure*}[t]
\centering
\includegraphics[width=\textwidth]{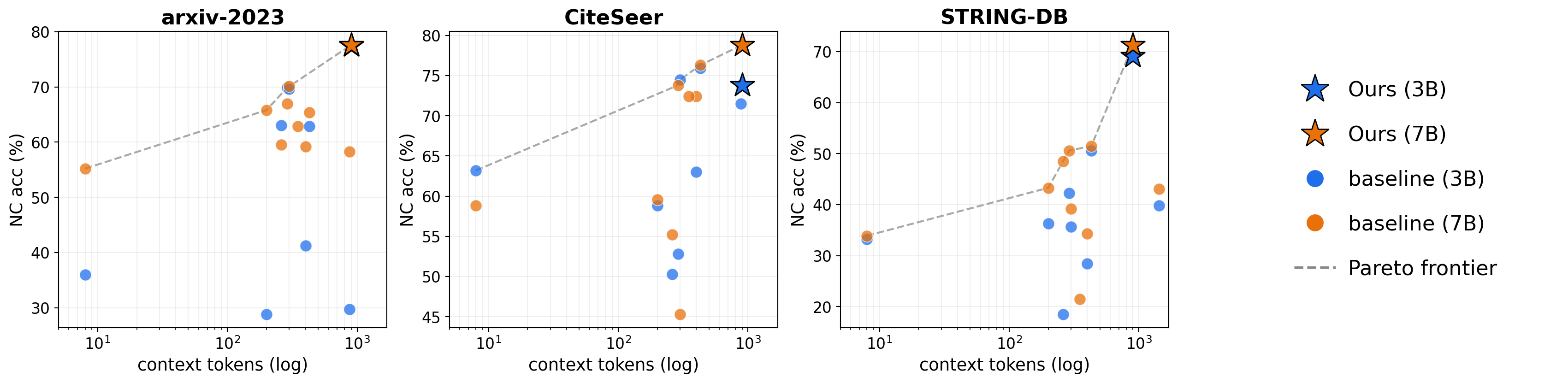}
\caption{Accuracy against inference context (part 2 of 2): arXiv-2023, CiteSeer,
STRING-db. Same conventions as Figure~\ref{fig:pareto-perdataset}: blue = 3B, orange = 7B;
star = ours, circle = baseline.}
\label{fig:pareto-perdataset-b}
\end{figure*}

A method is only useful if it reaches an answer at a reasonable cost, so we ask how
much context each approach spends per node: the average number of tokens the model
processes for one query, counting the prompt, every injected graph-token or
retrieved-text block, and the generated reasoning. For our method and the agentic
and static-token baselines we tokenise the recorded trajectories with the Qwen
tokeniser; for the single-pass graph-token baselines we read the budget off their
architecture. Figures~\ref{fig:pareto-perdataset} and~\ref{fig:pareto-perdataset-b} place
each method on the accuracy versus context plane for every dataset and both backbone sizes. The vertical position
is a method's measured node-classification accuracy from the main table; the
horizontal position is its per-query context.

For AgentGL we measure the context per dataset, since the amount of text it retrieves
varies with the corpus: its budget ranges from about $790$ tokens on Reddit to about
$1{,}430$ on STRING-db, where the protein descriptions are long, and on the heaviest
individual queries it passes $1{,}600$. Our own budget stays near $900$ tokens on
every dataset because each action contributes the same fixed-size block regardless of
the graph view it summarises, so the graph evidence never inflates the context the way
retrieved text does. The single-pass graph-token baselines are cheaper still, near
$300$ tokens, but their accuracy settles several points lower because one static block
cannot be revised as the reasoning proceeds. On every dataset except the homophilous
CiteSeer, where LLaGA leads at the 3B scale, our method occupies the accuracy-maximal
position on the frontier while keeping this bounded budget.

\section{Training-time analysis}
\label{app:time}

\begin{figure}[t]
\centering
\includegraphics[width=\columnwidth]{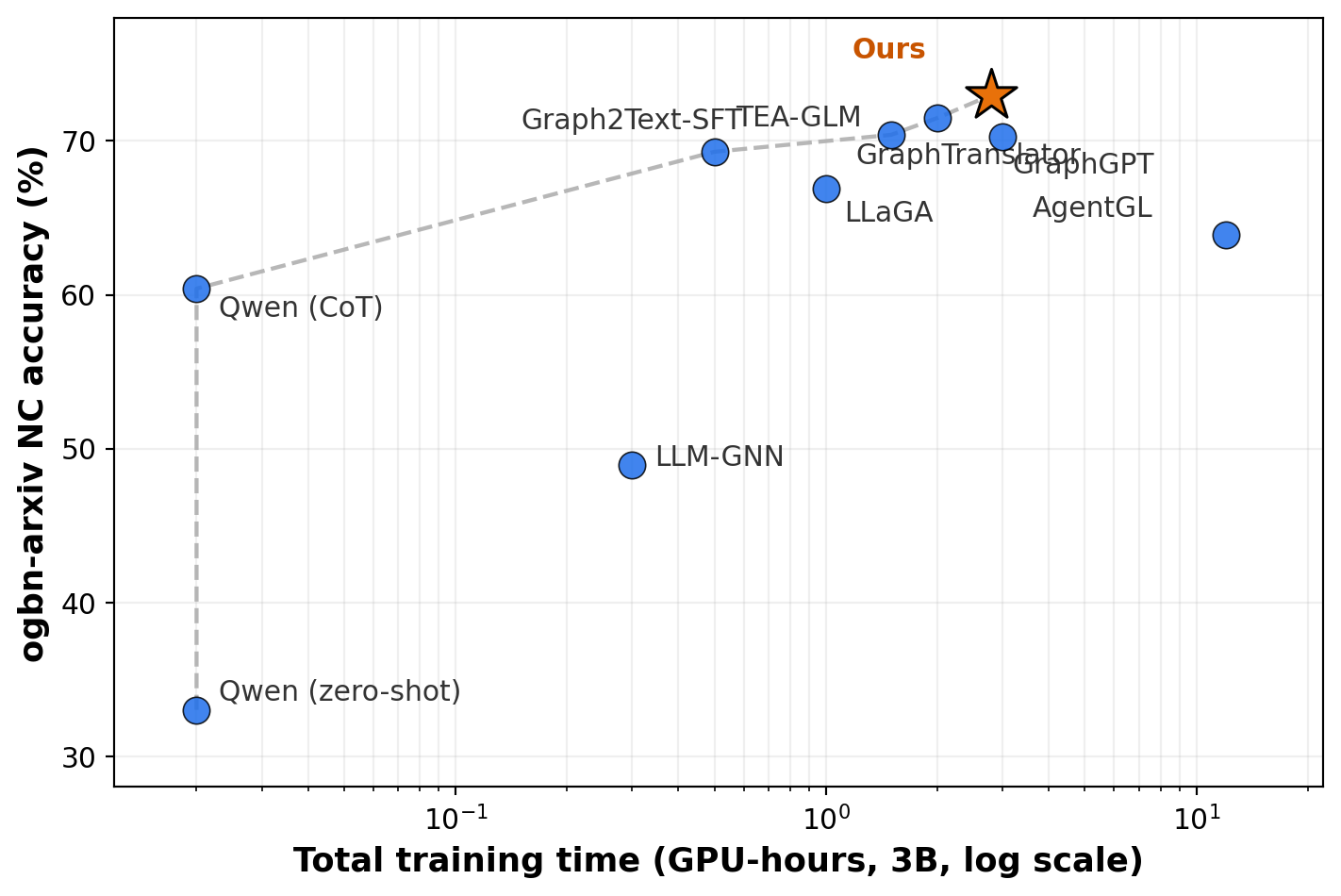}
\caption{Accuracy against total training time (3B configuration) on ogbn-arxiv. Our
three-stage pipeline (star) reaches the highest accuracy at a moderate, SFT-level budget
($\sim\!2.8$ GPU-hours), roughly $4\times$ below the RL-trained AgentGL.
Training-free baselines (zero-shot, chain-of-thought) are cheapest but lower in accuracy.
Times are single-GPU estimates gauged from each method's configuration and a measured
per-step throughput.}
\label{fig:time}
\end{figure}

\begin{figure}[t]
\centering
\includegraphics[width=\columnwidth]{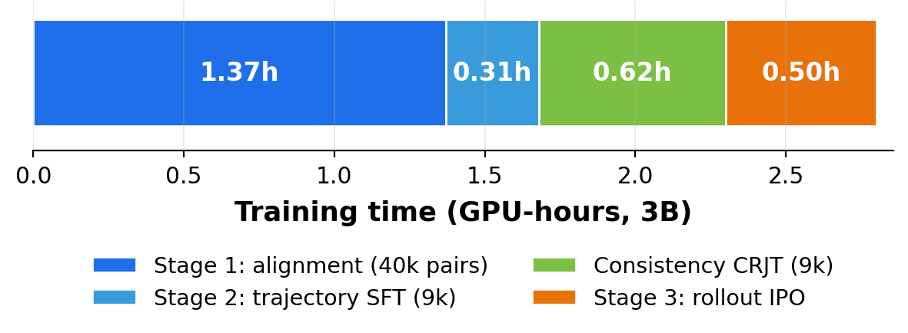}
\caption{Per-stage training-time breakdown of our three-stage 3B pipeline (Stage~2 shown
as its trajectory-SFT and consistency phases). The Stage~1
alignment pass over $40$k pairs is the largest single cost; the trajectory SFT,
consistency regularisation, and rollout IPO phases are comparatively cheap, for a total
of about $2.8$ GPU-hours.}
\label{fig:time-breakdown}
\end{figure}

Beyond inference context (\textbf{Appendix~\ref{app:pareto-perdataset}}), we compare methods by
their one-time training cost. Figure~\ref{fig:time} plots ogbn-arxiv node-classification
accuracy against total training time for the 3B configuration. We gauge each method on a
single GPU: for our pipeline and the fine-tuned baselines we multiply the number of
optimiser steps (corpus size times epochs, over the effective batch) by a measured
throughput of about two optimiser steps per second for the 3B backbone; the
reinforcement-learning baseline AgentGL is dominated by on-policy rollouts and is scaled up
per its rollout-heavy recipe; the zero-shot and chain-of-thought baselines need no training.
Our full pipeline (Stage~1 alignment on $40$k pairs, Stage~2 trajectory SFT plus
consistency regularisation on $9$k, and Stage~3 rollout IPO) totals about $2.8$ GPU-hours.
This sits in the same band as the single-stage fine-tuned baselines and roughly $4\times$ below the RL-trained AgentGL ($\sim\!12$ GPU-hours), while reaching the highest
accuracy. The training-free baselines are cheapest but trail in accuracy, chain-of-thought
being the strongest among them. Our method thus occupies the accuracy-maximal corner at a
moderate, supervised-fine-tuning-level budget. These are single-GPU estimates gauged from
each method's configuration and measured per-step throughput, intended as
order-of-magnitude comparisons rather than exact wall-clock. Figure~\ref{fig:time-breakdown}
breaks our budget down by stage: the Stage~1 alignment pass over $40$k pairs is the largest
single item at about $1.4$ hours, with the trajectory SFT, consistency, and rollout phases
together accounting for the remaining $1.4$ hours.

\section{Sensitivity analysis}
\label{app:sensitivity}

We study how two pooler design choices affect end-to-end accuracy: the number of tokens each
graph-token action emits and the GNN backbone inside the pooler. For each setting we run the
full pipeline with the $3$B backbone and report node-classification accuracy on the same
$1{,}000$-node test split, across four datasets: two citation graphs (ogbn-arxiv and
PubMed), a co-purchase graph (ogbn-products), and a social graph (Reddit).

\begin{figure}[t]
\centering
\includegraphics[width=\columnwidth]{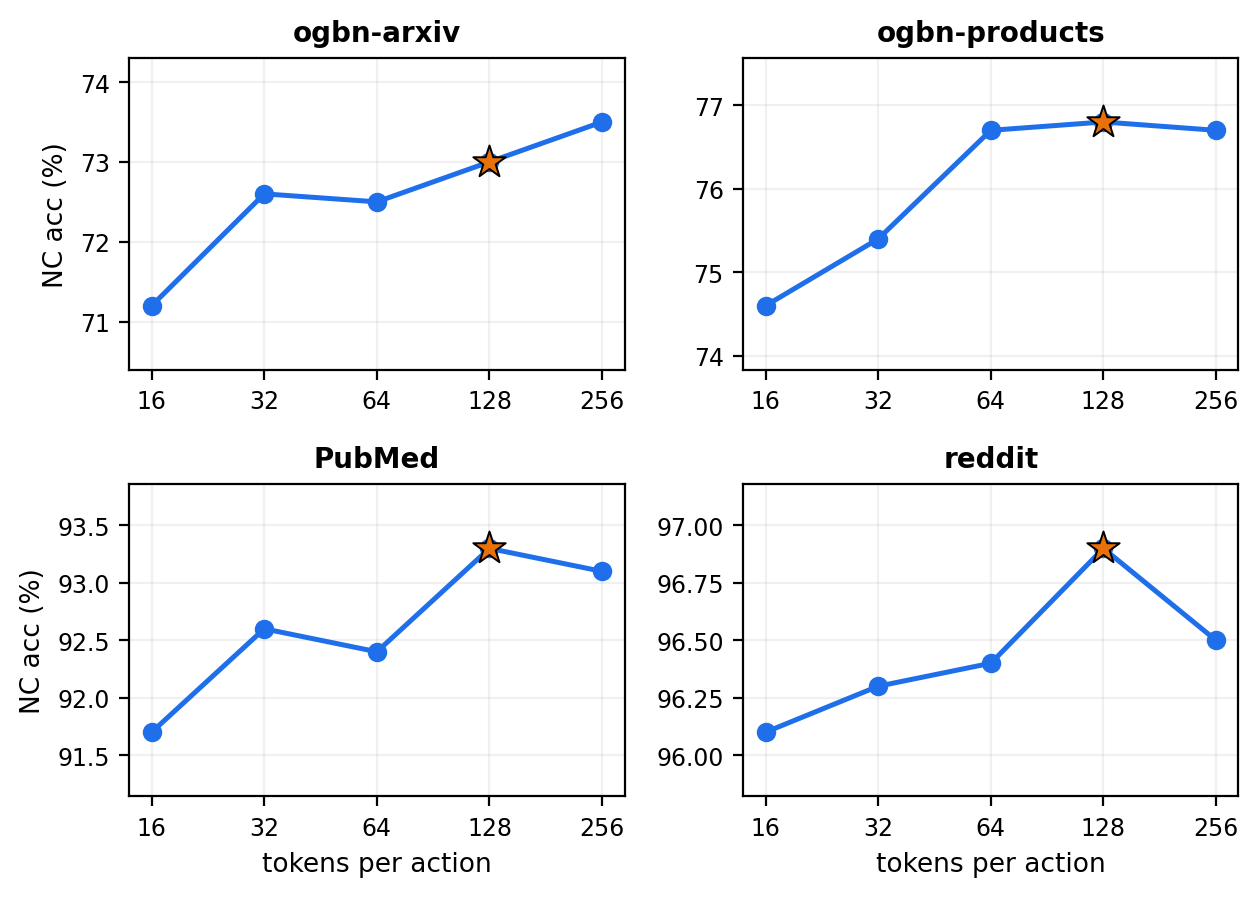}
\caption{Per-action token budget vs.\ full-pipeline NC accuracy ($3$B) on four datasets
(note the independent $y$-axis per panel). Accuracy saturates by roughly $64$--$128$ tokens;
the starred point marks the $128$-token default.}
\label{fig:sens-tokens}
\end{figure}

Figure~\ref{fig:sens-tokens} varies the per-action token budget. Accuracy rises steeply from
$16$ to $32$ tokens and then flattens: beyond $64$ tokens the four datasets move by only a few
tenths of a point, and $128$ is at or within noise of the best setting everywhere while
\textsc{ogbn-arxiv} still gains slightly at $256$. We use $128$ as the default, trading a
negligible accuracy change for half the inference context of $256$.

\begin{figure}[t]
\centering
\includegraphics[width=\columnwidth]{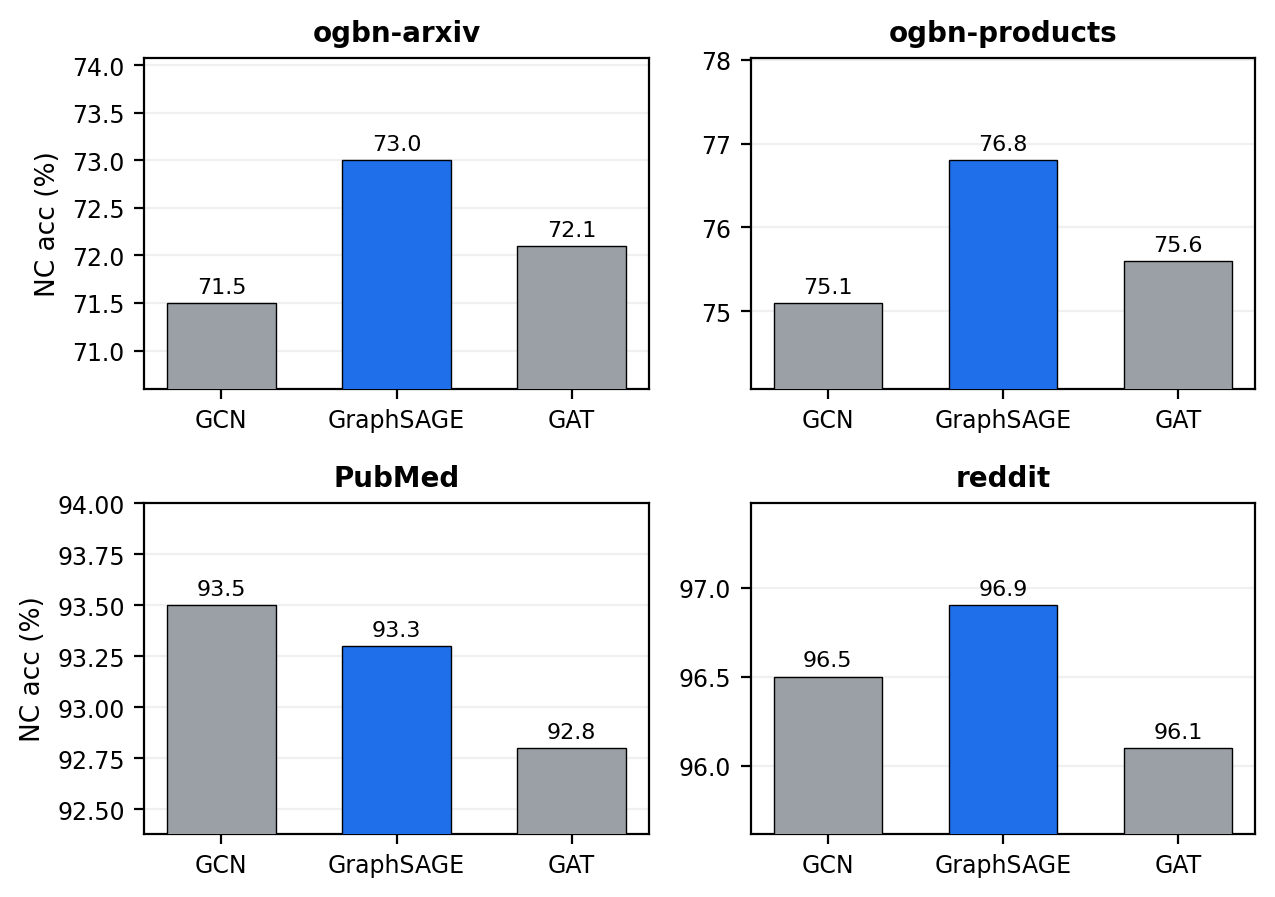}
\caption{GNN backbone vs.\ full-pipeline NC accuracy ($3$B, $128$ tokens per action; note the
independent $y$-axis per panel). The three convolutions stay within about a point and a half;
GraphSAGE (blue) is the default.}
\label{fig:sens-gnn}
\end{figure}

Figure~\ref{fig:sens-gnn} varies the GNN backbone at the default $128$ tokens. The pooler is
robust to this choice, with the three convolutions within about a point and a half on every
dataset. GraphSAGE is the strongest or tied-strongest on three of the four datasets and the
cheapest of the three, so we adopt it throughout.

\section{Trajectory case studies}
\label{app:case-studies}

The number of actions the policy issues is not fixed. Across datasets a trajectory
runs from a single action to the full ladder, and the length grows with how much the
node's own text leaves unresolved. We walk through four correct predictions, one at
each observed length.

\paragraph{One action.} A \textsc{Reddit} post reading ``recently got into japanese
knives and this 120mm wa-petty is my favorite'' (gold \texttt{chefknives}) is
unambiguous from its text alone. The policy issues a single \verb|anchor_text| echo of
the post and stops; no graph token is needed.

\paragraph{Two actions.} A \textsc{PubMed} article, ``altered glycolytic and oxidative
capacities of skeletal muscle contribute to insulin resistance'' (gold \texttt{Type 2
diabetes}), is clearly metabolic but its label is one of a few closely related diabetes
classes. The policy reads the anchor text and then a \verb|cluster_token| for the
community it sits in, whose members pin down the specific class.

\paragraph{Three actions.} An \textsc{ogbn-products} item, ``Brother high-speed label
printer with wireless networking'' (gold \texttt{Office Products}), could as easily be
filed under electronics. The policy escalates: \verb|anchor_text|, then a
\verb|cluster_token| for the product community, then a \verb|one_hop_token| for the item's
co-purchase neighbourhood; the co-purchased office equipment settles it on
\texttt{Office Products} instead of a generic electronics label.

\begin{figure*}[t]
\centering
\begin{tikzpicture}[
  font=\small,
  box/.style={rectangle, rounded corners, draw=black, thick, align=left, text width=4.3cm, inner sep=5pt},
  gtok/.style={box, fill=orange!14},
  txt/.style={box, fill=blue!9},
  io/.style={rectangle, rounded corners, draw=black, very thick, align=center, text width=11.5cm, inner sep=5pt, fill=gray!12},
  read/.style={rectangle, rounded corners, draw=gray!65, dashed, align=left, text width=7.4cm, inner sep=5pt, fill=gray!5},
  arr/.style={-{Latex[length=2.4mm]}, thick},
  rarr/.style={-{Latex[length=1.8mm]}, gray!60, thick},
]
\node[io] (in) {\textbf{Query paper:} ``gimme signals: discriminative signal encoding for multimodal activity recognition'' \quad gold label \texttt{cs.CV}\\[1pt] \footnotesize the text alone reads like signal processing};
\node[txt, below=0.5cm of in.south west, anchor=north west] (a1) {\textbf{1. anchor\_text}\\\footnotesize echo the paper's own title and abstract};
\node[gtok, below=0.42cm of a1] (a2) {\textbf{2. cluster\_token}\\\footnotesize pooled community token};
\node[gtok, below=0.42cm of a2] (a3) {\textbf{3. three\_hop\_token}\\\footnotesize pooled 3-hop neighbourhood token};
\node[gtok, below=0.42cm of a3] (a4) {\textbf{4. cluster\_token}\\\footnotesize re-encode the community};
\node[io, below=0.5cm of a4.south west, anchor=north west] (out) {predicted label: \texttt{cs.CV} \footnotesize (computer vision, not signal processing)};
\draw[arr] (a1.north |- in.south) -- (a1.north);
\draw[arr] (a1) -- (a2);
\draw[arr] (a2) -- (a3);
\draw[arr] (a3) -- (a4);
\draw[arr] (a4.south) -- (a4.south |- out.north);
\node[read, right=1.1cm of a2] (r2) {\textit{Stage-1 reader decodes this token as:}\\ ``deep learning for image classification with multiple labels''};
\node[read, right=1.1cm of a3] (r3) {\textit{Stage-1 reader decodes this token as:}\\ ``deep learning for image captioning''};
\node[read, right=1.1cm of a4] (r4) {\textit{Stage-1 reader decodes this token as:}\\ ``deep learning for image classification'' (same community, re-encoded)};
\draw[rarr] (a2.east) -- (r2.west);
\draw[rarr] (a3.east) -- (r3.west);
\draw[rarr] (a4.east) -- (r4.west);
\end{tikzpicture}
\caption{Agentic trajectory for one \textsc{ogbn-arxiv} node and what a Stage-1 reader
recovers from each injected graph token. The policy escalates from the paper's own text to
pooled community and neighbourhood tokens; decoding those tokens back to text shows they
carry computer-vision content (the discriminative signal for \texttt{cs.CV}),
recovering the research area of that neighbourhood, not a verbatim copy.}
\label{fig:trajectory-readout}
\end{figure*}
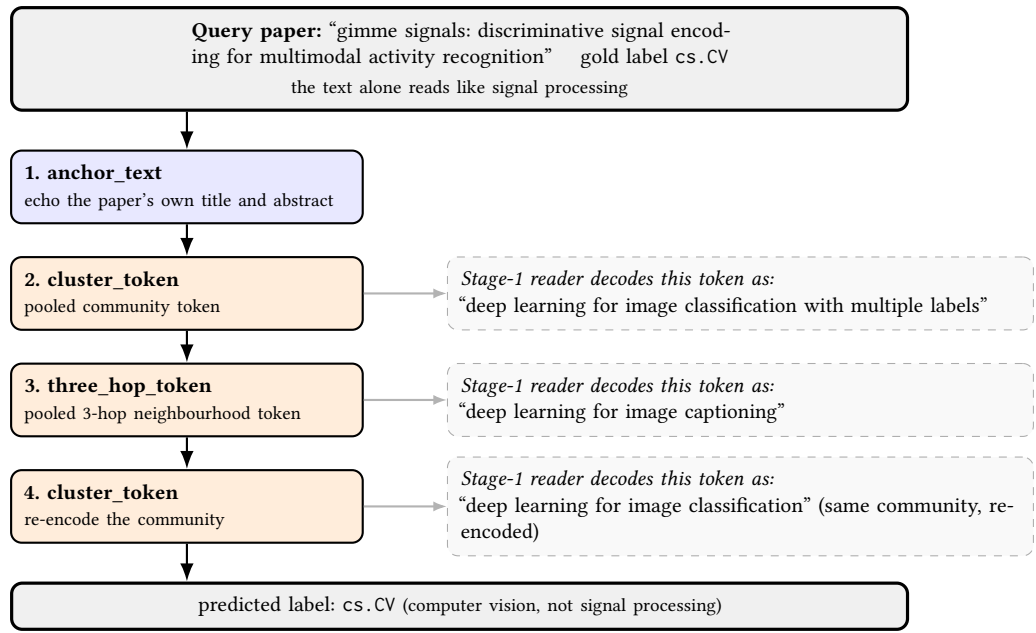

\paragraph{Four actions (Figure~\ref{fig:trajectory-readout}).} The longest of our case
studies is the \textsc{ogbn-arxiv} paper ``gimme signals: discriminative signal encoding for
multimodal activity recognition'' (gold \texttt{cs.CV}), whose title alone reads like signal
processing. Figure~\ref{fig:trajectory-readout} traces the four steps. The policy first
echoes the abstract with \verb|anchor_text|, which leaves the label unresolved. It then
issues a \verb|cluster_token| for the paper's community; passed through the Stage-1 reader
(below), that token decodes to ``deep learning for image classification with multiple
labels.'' Still not settled, it widens to a \verb|three_hop_token|, whose token reads
back as ``deep learning for image captioning,'' and closes with a second
\verb|cluster_token| that again decodes to image-classification content. All three graph
tokens land in computer-vision methodology, so the model answers \texttt{cs.CV} rather than
a signal-processing label.

\paragraph{Reading the injected tokens.} To check what the policy actually recovers
from a graph token, we decode it with a Stage~1 reader, a copy of the aligned model
asked to describe a node from its pooled token alone (using the Stage~1 encoder, not the
preference-tuned one). For the arxiv paper above (Figure~\ref{fig:trajectory-readout}), its
\verb|three_hop_token| and \verb|cluster_token| decode to computer-vision text, for
instance ``deep learning for image classification'' and ``image captioning,'' none of which
echoes the paper's own signal-processing wording. What the reader recovers is the research
area of the encoded graph view, not a verbatim copy, and this topical signal is what
pulls the prediction toward \texttt{cs.CV}. The behaviour is consistent across nodes: an
\textsc{albert} language-model paper decodes to text-generation content and a
knowledge-graph-embedding paper to machine-learning methodology, each landing in the right
area without reproducing the title. The pooled token thus carries enough topical structure
to place the node, though not enough to reconstruct its text.

In each case the policy spends more graph tokens exactly when the text is least
sufficient, and the graph view it chooses to encode carries the discriminative signal.

\end{document}